\documentclass[11pt, a4paper, logo, copyright, nonumbering]{map}
\usepackage{CJKutf8}
\usepackage{xargs}  
\usepackage{todonotes}
\usepackage{amsmath}
\usepackage{dsfont}

\usepackage{longtable}

\usepackage{svg}
\usepackage{fontawesome}
\usepackage{array}
\usepackage{tabularx}
\usepackage{latexsym}
\usepackage{graphicx}
\usepackage[utf8]{inputenc} 
\usepackage[utf8]{inputenc} 
\usepackage{amssymb} 
\usepackage{url}            
\usepackage{amsfonts}       
\usepackage{nicefrac}       
\usepackage{microtype}      
\usepackage{xspace}

\usepackage{listings}
\usepackage{makecell}
\usepackage{inconsolata}
\usepackage{multicol}
\usepackage{amstext}

\definecolor{darkblue}{RGB}{84, 112, 198}

\definecolor{lightblue}{rgb}{0.85, 0.95, 1.0}    
\definecolor{lightgreen}{rgb}{0.90, 1.0, 0.90}    
\definecolor{lightorange}{rgb}{1.0, 0.95, 0.85}   
\definecolor{lightpurple}{rgb}{0.95, 0.90, 1.0}   
\definecolor{lightgray}{rgb}{0.97, 0.97, 0.97}    

\usepackage{tikz}
\usetikzlibrary{calc}
\definecolor{battery-empty}{rgb}{0.9, 0.9, 0.9}
\newcommand{\difficultybar}[1]{%
  \begin{tikzpicture}[baseline, scale=0.5, every node/.style={scale=0.8}]
    \foreach \i in {1,2,3,4,5} {
      \ifnum\i>#1
        \draw[fill=battery-empty] (\i*0.5-0.5, 0) rectangle (\i*0.5, 0.25);
      \else
        \pgfmathsetmacro{\colorlevel}{80 - 12*(\i)} 
        \edef\x{\noexpand\draw[fill=blue!\colorlevel!white, opacity=0.9] (\i*0.5-0.5, 0) rectangle (\i*0.5, 0.25);}
        \x
        \draw[blue!50!black] (\i*0.5-0.5, 0) rectangle (\i*0.5, 0.25);
      \fi
    }
    \fill[battery-empty!70] (2.5, 0.08) rectangle (2.6, 0.17);
    \draw[battery-empty!70!black] (2.5, 0.08) rectangle (2.6, 0.17);
  \end{tikzpicture}%
}

\definecolor{hidden-draw}{RGB}{20,68,106}
\definecolor{hidden-pink}{RGB}{255,245,247}
\usepackage[edges]{forest}

\usepackage{bm}

\usepackage{natbib}
\usepackage{CJKutf8}
\usepackage{xargs}
\usepackage{todonotes}
\usepackage{multirow}
\usepackage{cleveref}
\usepackage{subcaption}
\usepackage{url}
\usepackage{svg}
\usepackage{fontawesome}
\usepackage{xcolor}
\usepackage{float}

\usepackage[utf8]{inputenc}
\usepackage{booktabs}
\usepackage{graphicx}
\usepackage{array}
\usepackage{tabularx}
\usepackage{xcolor,colortbl}  
\definecolor{boxcolor}{HTML}{d92523} 
\definecolor{bulbcolor}{HTML}{e3b87f} 
\usepackage{url}
\usepackage{ragged2e}   
\usepackage{makecell} 

\usepackage{hyperref}       
\usepackage{url}            
\usepackage{booktabs}       
\usepackage{nicefrac}       
\usepackage{microtype}      
\usepackage{xcolor}         
\usepackage{graphicx}
\usepackage{longtable}
\usepackage{array}
\usepackage{pbox}
\usepackage{tabularx}
\usepackage{multirow}
\usepackage{wrapfig}
\usepackage{pifont}
\usepackage{lipsum}
\usepackage{subcaption}
\usepackage{enumitem}
\usepackage{tikz}
\usepackage{xstring}

\usepackage{color-edits}

\usepackage{booktabs}
\usepackage{multirow}
\usepackage{tabularx}

\usepackage{parskip} 

\usepackage{threeparttable} 

\usepackage{xspace}
\usepackage{times}
\usepackage{latexsym}
\usepackage{booktabs}
\usepackage{fvextra}
\usepackage{graphicx}
\usepackage{subcaption} 
\usepackage[T1]{fontenc}
\usepackage{xcolor}
\usepackage[utf8]{inputenc}

\usepackage{microtype}

\usepackage{inconsolata}
\usepackage{amsmath}
\usepackage{booktabs}
\usepackage{graphicx}
\usepackage{pifont}
\usepackage{multirow}          
\usepackage{colortbl}          
\usepackage[linesnumbered,ruled,vlined]{algorithm2e}
\usepackage{xcolor}
\usepackage{amssymb}

\definecolor{rliableolive}{HTML}{BBCC33}
\definecolor{rliableblue}{HTML}{77AADD}
\definecolor{rliablered}{HTML}{f63c44}
\definecolor{darkgreen}{RGB}{0, 100, 0}

\newcommand{\ie}{\textit{i.e.,}\xspace}

\newcommand{\eg}{\textit{e.g.,}\xspace}

\newcommand{\paratitle}[1]{\vspace{1.5ex}\noindent\textbf{#1}}

\usepackage{alltt}
\usepackage[most,skins,theorems]{tcolorbox}
\tcbuselibrary{skins,breakable,fitting,hooks}  
\tcbset{
  aibox/.style={
    width=\linewidth,
    top=7pt,
    bottom=8pt,
    colback=white,
    colframe=black,
    colbacktitle=black,
    enhanced,
    center,
    attach boxed title to top left={yshift=-0.1in,xshift=0.15in},
    boxed title style={boxrule=0pt,colframe=white},
  }
}
\tcbset{
  aibox2/.style={
    width=\linewidth,
    top=7pt,
    bottom=2pt,
    colback=green!18!white,
    colframe=black,
    colbacktitle=black,
    enhanced,
    center,
    attach boxed title to top left={yshift=-0.1in,xshift=0.15in},
    boxed title style={boxrule=0pt,colframe=white,},
  }
}
\newtcolorbox{AIbox}[2][]{aibox,title=#2,#1}
\newtcolorbox{AIbox2}[2][]{aibox2,title=#2,#1}

\usepackage{xcolor}

\hypersetup{
    colorlinks=true,            
    linkcolor=blue,             
    filecolor=magenta,          
    urlcolor=cyan,              
    citecolor=purple,             
    pdftitle={Overleaf Example},
    pdfpagemode=FullScreen,
}

\definecolor{iquestblue}{HTML}{173C7F}
\definecolor{iquestazure}{HTML}{528FCC}

\newcommandx{\info}[2][1=]{\todo[linecolor=red,backgroundcolor=red!25,bordercolor=red,#1]{#2}}

\title{
\vspace{-0.2in}
\centering \fontsize{15pt}{16pt}\selectfont
Spectral-Driven Adaptive-Rank Optimizer for Large Language Model Reinforcement Learning with Verifiable Rewards
\vspace{-0.2in}
}

\author{
Zhipeng Chen\textsuperscript{1},
Wayne Xin Zhao\textsuperscript{1$\dagger$},
\normalfont
Ji-Rong Wen\textsuperscript{1}\\
\normalfont
\textsuperscript{1}Gaoling School of Artificial Intelligence, Renmin University of China, \\
\normalfont
\normalsize{
\textsuperscript{$\dagger$}Corresponding Authors \\
\normalfont
Email: \texttt{zhipeng\_chen@ruc.edu.cn},\quad
\texttt{batmanfly@gmail.com}
}}

\newcommand{\ignore}[1]{}

\usepackage[most]{tcolorbox}
\definecolor{darkorange}{RGB}{255, 140, 0}
\definecolor{lightgreen}{RGB}{145, 204, 117}
\definecolor{lightyellow}{RGB}{250, 200, 88}
\definecolor{lightred}{RGB}{238, 102, 102}
\definecolor{lightblue}{RGB}{115, 192, 222}

\definecolor{gray_1}{HTML}{B7B7B7}
\definecolor{gray_2}{HTML}{F0F0F0} 
\definecolor{frame_blue}{HTML}{A9D18E}

\newtcolorbox[auto counter, number within=section]{PromptBox}[2][]{
    enhanced,
    breakable,
    colback=gray_2, 
    colframe=gray_1,
    coltitle=white,
    fontupper=\small,
    fonttitle=\bfseries,
    title={#2}, 
    label={#1},
    arc=2pt,
    boxrule=1pt,
    left=2mm, right=2mm, top=2mm, bottom=2mm,
}

\usepackage{etoolbox}
\makeatletter
\patchcmd{\@maketitle}
  {\@toptitlebar}
  {%
    \hbox to \textwidth{%
    
    \includegraphics[height=0.7cm]{figures/aibox-logo.png}%
      \hfill
      \includegraphics[height=0.7cm]{figures/logo2.pdf}%
    }%
    \vspace{0.3em}
    \@toptitlebar
  }
  {}{}
\makeatother

\begin{abstract}

Reinforcement learning with verifiable rewards (RLVR) is reshaping how large language models (LLMs) acquire reasoning capabilities, yet the optimizers used remain largely inherited from teacher-forcing training. 
This raises a fundamental question: does RLVR induce a distinct optimization geometry that calls for a different optimizer?
We investigate this question through the spectral structure of LLM gradients
and find that gradient energy progressively concentrates into a shrinking effective subspace, which is characteristic of RLVR, as it is absent under supervised fine-tuning (SFT) and is largely removed when the verifiable reward is replaced by a random one.
Moreover, the rank required to capture a fixed fraction of gradient energy steadily decreases over training, revealing that the effective dimensionality of useful updates itself evolves with training. 
These findings motivate a new optimization principle: rather than prescribing the update rank externally, an optimizer should adapt its rank to the evolving spectral structure of the gradient.
We introduce \textbf{Ares} (\textbf{A}daptive \textbf{R}ank-\textbf{E}ffective \textbf{S}pectral Optimization), which estimates the effective update rank from the gradient's spectral energy distribution, tracks its evolution, and confines the gradient signal driving each update to the resulting low-rank subspace.
Across LLMs of different scales, Ares improves accuracy over widely used optimizers and generalizes across RLVR algorithms and tasks. 
Results suggest that effective optimization may require optimizers to be designed around the geometry and dynamics of the training process.
We release the code and data at \url{https://github.com/RUCAIBox/Ares}.

\end{abstract}

\begin{document}
\maketitle

\let\oldthefootnote\thefootnote



\section{Introduction}
\label{sec:intro}

Large language models (LLMs) are increasingly optimized via reinforcement learning with verifiable rewards (RLVR), a central component of training strong reasoning models~\citep{deepseekr1,o1}.
Yet the optimizers driving this stage, from AdamW~\citep{loshchilov2019decoupled} to preconditioning and orthogonalization methods~\citep{gupta2018shampoo,jordan2024muon}, are largely inherited from supervised training and carried into RLVR essentially unchanged.
Rather than assuming that RLVR inherits the optimization geometry of supervised training, we uncover its distinctive gradient geometry and identify the key property that should guide optimizer design.

Why might these updates differ? Unlike teacher forcing, whose signal is anchored to reference continuations, RLVR aggregates signed contributions from sampled trajectories: positive-advantage samples raise their likelihood while negative-advantage ones lower it~\citep{shao2024deepseekmath}. 
Because this signal is drawn from the policy's own evolving distribution, the update direction depends on sampling and shifts during training, and it typically acts on an already capable policy rather than one still forming its representations. 
These differences raise an empirical question: does the distinct update dynamics of RLVR leave an observable imprint on the spectral structure of its gradients?

\begin{figure}[t]
    \centering
    \includegraphics[width=0.95\linewidth]{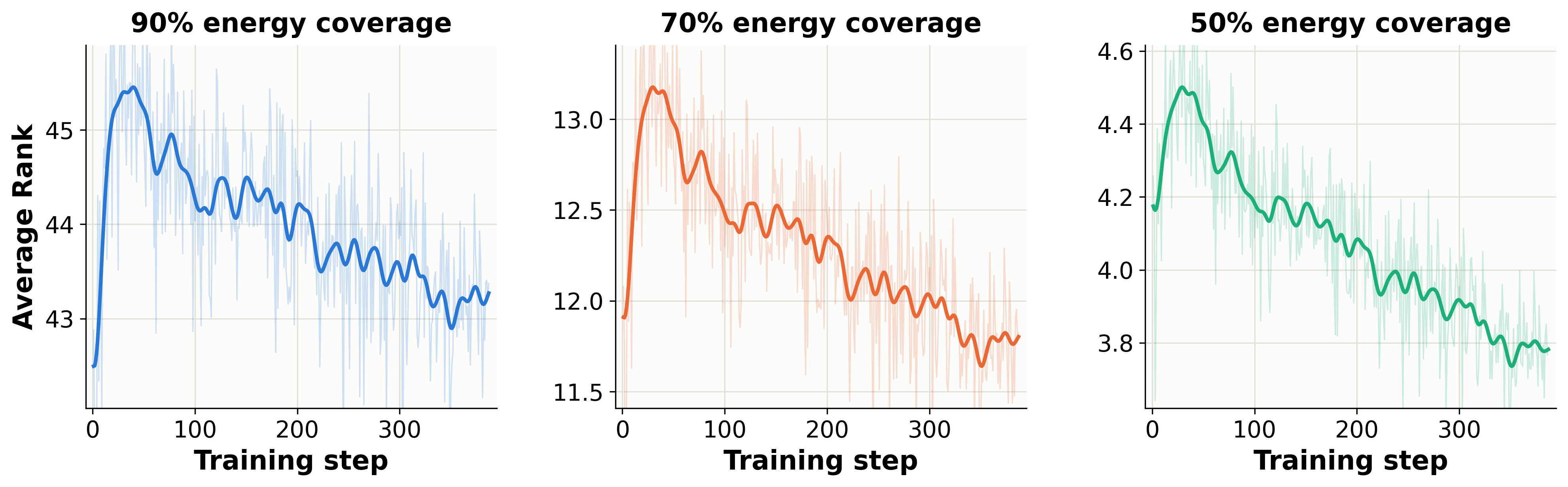}
    \caption{The rank $r(\varepsilon)$ needed to capture a fixed fraction $\varepsilon$ of gradient energy (Qwen3-1.7B, GRPO, AdamW). After a brief early transient, $r(\varepsilon)$ declines steadily. Curves are lightly smoothed.}
    \label{fig:rank_for_coverage}
\end{figure}

\begin{wrapfigure}{r}{0.4\linewidth}
    \includegraphics[width=\linewidth]{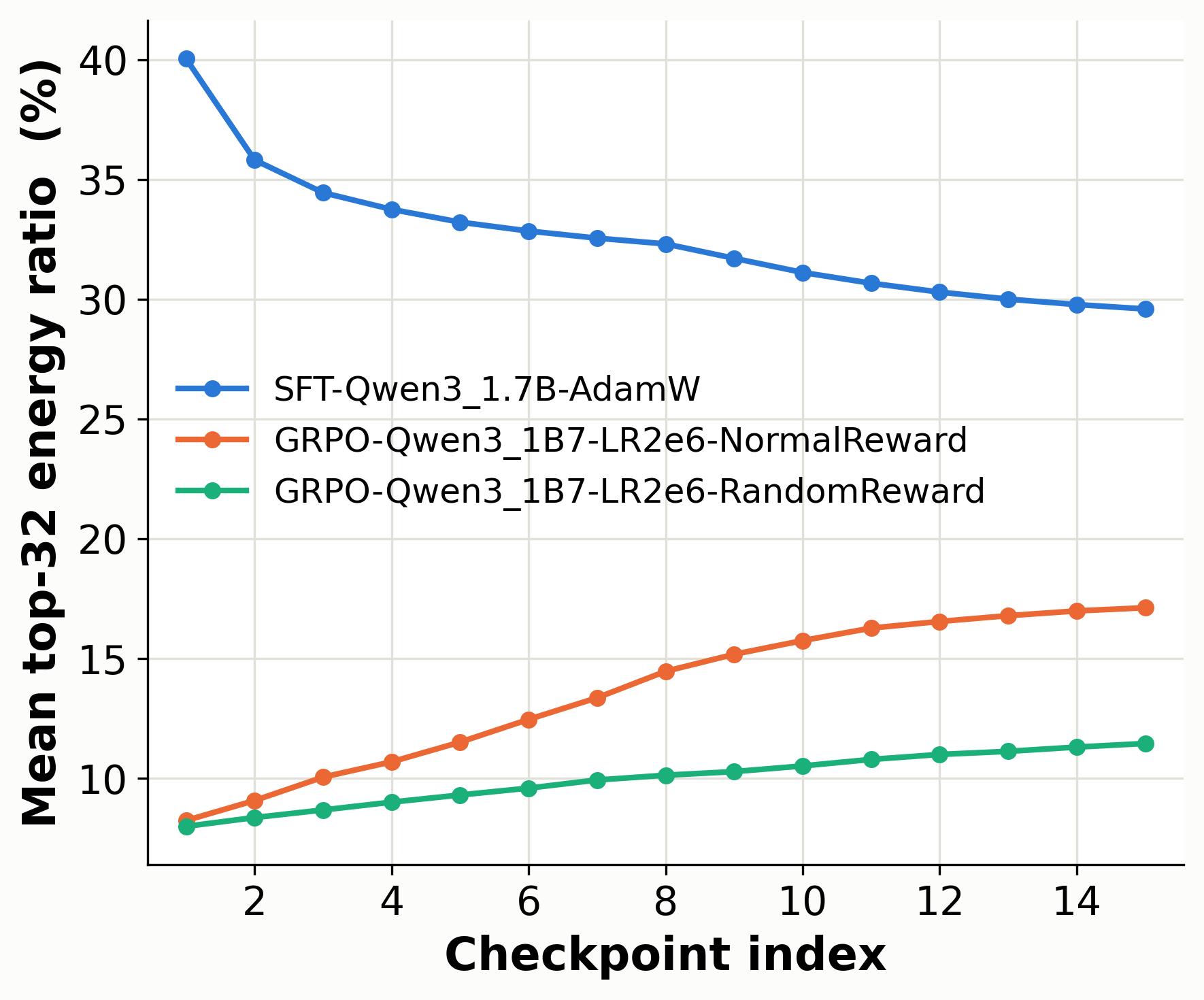}
    \caption{Top-32 energy ratio of gradient via RLVR (Qwen3-1.7B). We emphasize the trends, not absolute levels.}
    \label{fig:spectral_concentration}
\end{wrapfigure}

They do. Inspired by previous work~\cite{NExt}, we measure the spectral structure of LLM gradient matrices throughout RLVR with the energy captured by the leading singular components, which describes how concentrated the spectrum is, not the algebraic rank, which is almost always full for a dense gradient.
Across RLVR, gradient energy progressively concentrates in the leading singular components (Fig.~\ref{fig:spectral_concentration}).
In our experiments, this behavior is characteristic of RLVR rather than of fine-tuning in general: supervised fine-tuning (SFT) shows no such rise but a slight broadening, and replacing the reward with a random one, leaving rollouts and configuration intact, removes most of the increase, indicating that the trend is not explained solely by on-policy sampling.

This suggests that rank in RLVR is not merely a compression parameter but a quantity characterizing the evolving optimization geometry. 
Yet a concentrating spectrum need not imply a changing \emph{useful} rank, since a large enough fixed rank could cover the whole run. 
To test this, we track the rank $r(\epsilon)$ needed to capture a fixed fraction $\epsilon$ of gradient energy during RLVR (Fig.~\ref{fig:rank_for_coverage}). 
After a brief transient, $r(\epsilon)$ declines steadily.
The effective dimensionality of the update, not merely the shape of the spectrum, evolves during training.
Indeed, for RLVR gradients a small rank already suffices to cover the vast majority of the update subspace, and this coverage is attained at a shrinking rank as training proceeds.
This motivates an optimizer that infers its update rank from the observed spectrum rather than prescribing it externally.

Actually, existing optimizers do not exploit this. 
Low-rank methods such as GaLore~\citep{zhao2024galore} treat low rank as a compression device for memory-efficient pretraining, with an externally fixed rank; preconditioning and orthogonalization methods (\eg Shampoo~\citep{gupta2018shampoo} and Muon~\citep{jordan2024muon}) adapt during training but do not use the evolving spectral complexity of the gradient to set the dimensionality of the update subspace. 
The issue is thus not whether the gradient is low-rank, but where the rank comes from: a fixed small rank truncates non-negligible components when the spectrum broadens, an overly large one wastes capacity when it is sharply concentrated, and a constant or preset schedule sets the rank independently of the gradient. 
We present a detailed taxonomy of optimizer families on this axis in Sec.~\ref{sec:discussion} (Table~\ref{tab:rank_taxonomy}).

We introduce \textbf{Ares} (\textbf{A}daptive \textbf{R}ank-\textbf{E}ffective \textbf{S}pectral Optimization), which infers the update rank from the gradient's instantaneous spectral energy distribution, smoothly tracks its evolution, and confines the gradient signal driving each update to the resulting low-rank subspace, using randomized subspace iteration~\citep{halko2011finding} to avoid a full SVD.
Across LLMs at various scales, Ares improves average accuracy over widely used optimizers, with gains that hold across RLVR algorithms and extend beyond mathematics, and ablations confirm that each component contributes.
Our contributions are as follows,

$\bullet$ \textbf{Spectral characterization of RLVR gradients:} we identify a persistent concentration of gradient energy in the leading singular components during RLVR training, with a declining entropy-based effective rank, which we do not observe under SFT or random-reward controls in the same setup.

$\bullet$ \textbf{A spectral-driven adaptive optimizer:} we propose Ares, which estimates effective rank from gradient energy distributions and dynamically adjusts its update subspace during RLVR training, rather than fixing the rank or following a preset schedule.

$\bullet$ \textbf{Performance improvement:} we validate this spectral-driven principle across model scales, downstream tasks, and RLVR algorithms, and isolate the contributing components through ablations.

\section{Related Work}
\label{sec:related}

\paratitle{RLVR for LLMs.}
RLVR has become the dominant paradigm for post-training LLMs, first for aligning models with human preferences~\citep{ouyang2022instructgpt,schulman2017ppo}, and recently for eliciting strong reasoning through verifiable rewards~\citep{deepseekr1,lambert2024tulu3}. 
In this RLVR setting, a large body of work has focused on the policy-optimization objective. 
PPO~\citep{schulman2017ppo} relies on a learned value model, which GRPO~\citep{shao2024deepseekmath} removes by estimating advantages from sampled responses; subsequent variants such as REINFORCE++~\citep{hu2025reinforceplusplus} and DAPO~\citep{yu2025dapo} further simplify or stabilize this estimate. 
These methods reshape the training signal but leave the underlying optimizer untouched, typically inheriting AdamW~\citep{loshchilov2019decoupled} from SFT. 
In contrast, we study the optimization geometry that RLVR induces and design the optimizer around it.

\paratitle{Optimizers for LLM Training.}
Adam~\citep{kingma2015adam} and AdamW~\citep{loshchilov2019decoupled} remain the default optimizers for training LLMs. 
A large body of work seeks to improve upon them, either by exploiting curvature through matrix preconditioning (\eg Shampoo~\citep{gupta2018shampoo}, SOAP~\citep{vyas2024soap}, and the diagonal second-order method Sophia~\citep{liu2024sophia}) or by orthogonalizing the update, as in Muon~\citep{jordan2024muon}, while others discover update rules through symbolic search~\citep{chen2023lion}. 
A complementary line targets memory efficiency: Adafactor~\citep{shazeer2018adafactor} and CAME~\citep{luo2023came} factor the optimizer state, and GaLore~\citep{zhao2024galore} projects gradients onto a low-rank subspace, motivated by the fact that training progresses within a low-dimensional subspace~\citep{gurari2018gradient}. 
However, these methods treat rank as an externally fixed compression budget rather than a property of the gradient, and none is designed for RLVR.
Our optimizer infers the update rank from the evolving spectral structure of RLVR gradients, adapting the subspace dimension online throughout RLVR.

\section{Preliminary}
\label{sec:preliminary}

\paratitle{RLVR for LLMs.}
RLVR fine-tunes a language model policy $\pi_\theta$ so that it produces correct solutions to problems drawn from a data distribution $\mathcal{D}$. Given a prompt $x \sim \mathcal{D}$, the policy samples a response $y \sim \pi_\theta(\cdot \mid x)$, which is scored by a rule-based verifier $r(x, y) \in \{0, 1\}$ that returns $1$ only when $y$ reaches the correct final answer. The training objective is to maximize the expected reward while staying close to a reference policy $\pi_{\text{ref}}$:
\begin{equation}
\small
    \max_{\theta}\;
    \mathbb{E}_{x \sim \mathcal{D},\, y \sim \pi_\theta(\cdot \mid x)}\big[\, r(x, y)\,\big]- \beta\, \mathbb{D}_{\mathrm{KL}}\!\big(\pi_\theta(\cdot \mid x)\,\|\,\pi_{\text{ref}}(\cdot \mid x)\big),
\end{equation}
where $\beta$ controls the strength of the KL regularization. 
Since $r$ is non-differentiable, this objective is optimized with policy-gradient methods. 
In this work we mainly adopt a popular algorithm GRPO~\citep{shao2024deepseekmath}, which samples a group of $G$ responses $\{y_i\}_{i=1}^{G}$ per prompt and estimates a group-relative advantage $\hat{A}_i = (r_i - \mathrm{mean}(\{r_j\})) / \mathrm{std}(\{r_j\})$, avoiding a separately learned value model. 
The corresponding gradient reweights the log-likelihood of each sampled token by its advantage, $\nabla_\theta \mathcal{J} \propto \mathbb{E}\big[\hat{A}_i\, \nabla_\theta \log \pi_\theta(y_i \mid x)\big]$, so that positive-advantage responses are reinforced and negative ones suppressed.
Furthermore, importance sampling and clipping are employed to reduce the variance caused by off-policy sampling and enhance the stability of RLVR training.

\paratitle{AdamW Optimizer.}
Given the estimated gradient $g_t$ of the loss with respect to a parameter at step $t$, AdamW~\citep{loshchilov2019decoupled} maintains exponential moving averages of the first and second moments, $m_t = \beta_1 m_{t-1} + (1 - \beta_1) g_t$ and $v_t = \beta_2 v_{t-1} + (1 - \beta_2) g_t^2$, and applies a bias-corrected, per-coordinate adaptive update with decoupled weight decay,
\begin{equation}
\small
    \theta_t = \theta_{t-1}
    - \eta \left( \frac{\hat{m}_t}{\sqrt{\hat{v}_t} + \epsilon} + \lambda\, \theta_{t-1} \right),
\end{equation}
where $\hat{m}_t$ and $\hat{v}_t$ are the bias-corrected moments, $\eta$ is the learning rate, and $\lambda$ the weight-decay coefficient. 
AdamW is the default optimizer for language-model training and is inherited essentially unchanged into RLVR post-training, which serves as the primary baseline in this paper.

\section{Approach}
\label{sec:approach}

\begin{figure}[t]
    \includegraphics[width=\linewidth]{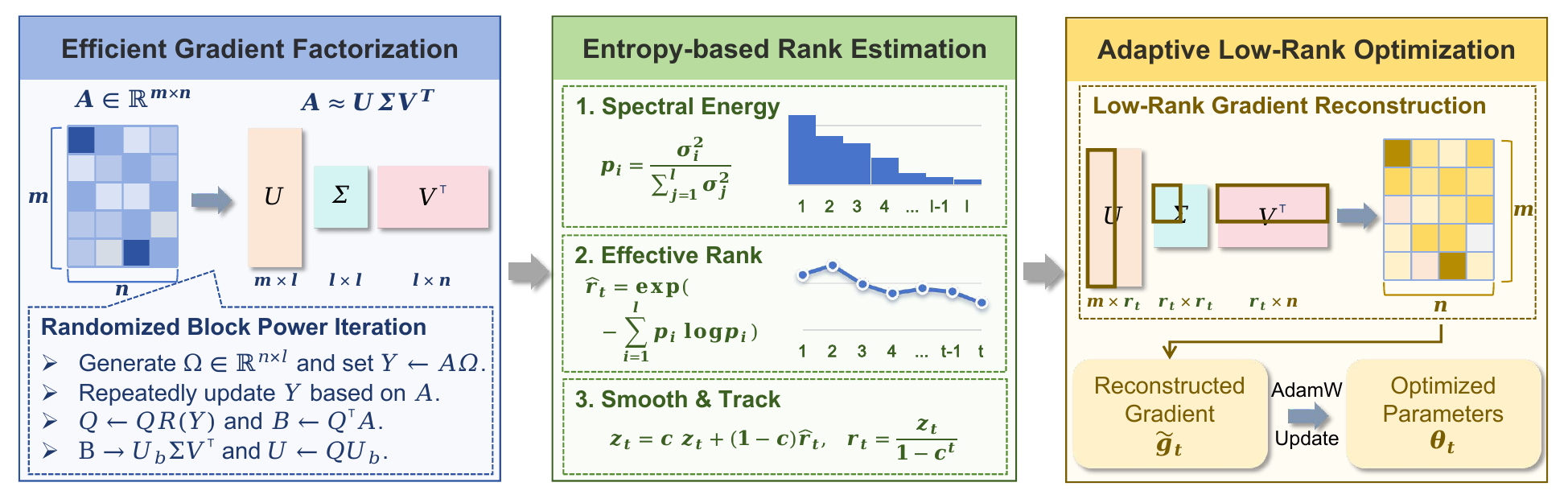}
    \caption{Ares overview. For 2-D tensors, Ares first decomposes the gradients using an approximate algorithm. Then, Ares estimates the effective rank based on the spectral energy and the rank momentum. Next, Ares uses the effective subspace to reconstruct the gradients, which is utilized to update parameters. 1-D tensors will be updated via standard AdamW. The pseudo code is in Appendix~\ref{app:pseudocode}.}
    \label{fig:overview}
\end{figure}


As RLVR gradients progressively concentrate in leading singular components, their effective update dimensionality declines throughout training, motivating \textbf{Ares} to adapt its update rank directly to the observed spectrum. 
At each step, Ares (1) factorizes the gradient to identify its leading singular subspace (Sec.~\ref{sec:factorization}), (2) estimates and smoothly tracks the effective rank (Sec.~\ref{sec:rank_estimation}), and (3) updates within the resulting low-rank subspace (Sec.~\ref{sec:lowrank_opt}). Fig.~\ref{fig:overview} illustrates the overall framework.

\subsection{Efficient Gradient Factorization}
\label{sec:factorization}


To estimate the effective rank and remove noise during RLVR, Ares only needs the leading singular subspace at each step, yet dense SVD computes the entire spectrum at a cost quadratic in the smaller dimension $n$. Since RLVR gradients concentrate most of their energy in a small number of leading components (Fig.~\ref{fig:rank_for_coverage}), resolving the low-energy tail is unnecessary. Ares therefore uses randomized block power iteration (RBPI)~\citep{halko2011finding} to extract the leading subspace directly, with cost governed by the target rank rather than the ambient dimension.

Given a gradient matrix $A\in\mathbb{R}^{m\times n}$, a target rank $k$, an oversampling parameter $p$, and a number of power iterations $q$, RBPI draws a Gaussian test matrix $\Omega\in\mathbb{R}^{n\times\ell}$ with block size $\ell=k+p$ and forms the sketch $Y=A\Omega$. It then alternately applies $A^\top$ and $A$ with a QR re-orthonormalization after each application, repeated $q$ times, so that the resulting block behaves like $(AA^\top)^q A\,\Omega$. Since $(AA^\top)^q A$ shares the singular vectors of $A$ but has singular values $\sigma_i^{2q+1}$, this power scheme amplifies the separation between the leading and trailing spectrum before the subspace is extracted. Orthonormalizing the block yields a basis $Q$, and projecting $A$ onto it gives the small matrix $B=Q^\top A$, whose SVD is cheap; composing its left factors with $Q$ recovers approximate singular vectors and values of $A$. The procedure outputs a factorization with left factor $U\in\mathbb{R}^{m\times\ell}$, singular values $\Sigma=\operatorname{diag}(\sigma_1,\dots,\sigma_\ell)$, and right factor $V\in\mathbb{R}^{n\times\ell}$ (so that $A\approx U\Sigma V^\top$), which Sec.~\ref{sec:rank_estimation} uses to estimate the effective rank.

This construction replaces the $O(mn^2)$ cost of a dense SVD with $O(q\,mn\ell)$, an advantage that is substantial precisely when $q(k+p)\ll n$, the regime Ares operates in since transformer weight gradients have $n$ on the order of $10^3$ to $10^4$ while the target rank is a few tens. The accuracy of the approximation is likewise favorable: RBPI is exact when $A$ is genuinely rank-$k$, and in general its projection error is bounded by a multiplicative factor times the optimal rank-$k$ error $\sigma_{k+1}$, with the factor tightening as $q$ grows and as the spectral gap widens. The spectral concentration of RLVR gradients established in Sec.~\ref{sec:intro} supplies exactly such a gap, so a moderate $q$ and a small oversample let RBPI track the leading subspace closely. The full cost and analysis is in Appendix~\ref{app:rbpi}.

\subsection{Spectral Energy-based Rank Estimation}
\label{sec:rank_estimation}

The factorization exposes the gradient spectrum, but it does not by itself specify how many components the update should retain. 
Sec.~\ref{sec:intro} shows that this number is neither constant across training nor well described by the algebraic rank, which is almost always full for a dense gradient; what evolves is the concentration of energy. 
Ares therefore reads the rank off the spectral energy distribution and tracks its evolution smoothly over training.

\paratitle{Spectral Energy Modeling.}
We characterize the spectrum by the energy carried by each singular direction rather than by the presence or absence of components.
Writing $\sigma_1\ge\cdots\ge\sigma_{k+p}\ge 0$ for the singular values returned by the factorization, we form the normalized energy distribution
\begin{equation}
\small
    p_i = \frac{\sigma_i^2}{\sum_{j} \sigma_j^2},
\end{equation}
so that $p_i$ measures the fraction of gradient energy along the $i$-th direction.
Since RBPI returns only the leading $\ell=k+p$ components, the normalization is over the retained subspace rather than the full gradient, so $\hat r_t$ below is an effective rank measured \emph{within the target-rank budget} $k$.
We then summarize the concentration of this distribution by its Shannon entropy and convert it into an effective rank~\citep{effective_rank} through
\begin{equation}
\small
    H_t = -\sum_{i} p_i \log p_i,
    \qquad
    \hat r_t = \exp(H_t).
\end{equation}
The $\hat r$ admits a direct interpretation as the number of directions over which the energy is effectively spread: when the energy is uniform across the $\ell$ retained directions the entropy is maximal and $\hat r_t$ attains its upper bound $\ell$, whereas when the energy collapses onto a few leading components the entropy shrinks and $\hat r_t$ becomes small.
As it depends on the instantaneous spectrum, $\hat r_t$ is a measure of the current spectral complexity of the gradient and moves as the spectrum evolves, which is precisely the behavior of the declining $r(\epsilon)$ in Fig.~\ref{fig:rank_for_coverage}.

\paratitle{Rank Adaptation.}
The per-step estimate $\hat r_t$ inherits the stochasticity of the sampled gradient, and using it directly would cause the dimension of the update subspace to fluctuate from step to step. Ares therefore smooths the estimate with an exponential moving average across steps. Let $c$ denote the smoothing coefficient and $z_t$ an uncorrected EMA accumulator initialized at $z_0=0$. We update
\begin{equation}
\small
    z_t = c\, z_{t-1} + (1-c)\, \hat r_t,
    \qquad
    r_t = \operatorname{clip}\!\left( \operatorname{round}\!\Big( \frac{z_t}{1 - c^{\,t}} \Big),\; 1,\; k \right),
\end{equation}
where the accumulator $z_t$ is kept separate from the tracked rank so that the division by $1-c^{\,t}$ implements the standard EMA bias correction from a zero start, $\operatorname{round}(\cdot)$ maps the real-valued effective rank to an integer index, and the clipping keeps the rank within $[1,k]$. 
Because $r_t$ can only decrease from the ceiling $k$, Ares performs adaptive truncation within a preset maximum-rank budget rather than discovering a rank above $k$.
The tracked rank $r_t$ thus follows the concentration of gradient energy continuously over training while remaining stable against per-step noise.
This is the sense in which the rank in Ares is inferred from the observed spectrum and tracked as it evolves, in contrast to a fixed rank as in GaLore, the distinction made explicit in Sec.~\ref{sec:discussion}. 

\subsection{Adaptive Low-Rank Optimization}
\label{sec:lowrank_opt}

Based on $r_t$, the final stage forms the parameter update within the selected subspace.
Ares reconstructs a low-rank gradient by truncating the factorization of Sec.~\ref{sec:factorization} to its leading $r_t$ components,
\begin{equation}
\small
    \tilde g_t = U_{:,\,1:r_t}\, \Sigma_{1:r_t}\, V^\top_{1:r_t,\,:},
\end{equation}
which is the rank-$r_t$ truncated SVD reconstruction: it projects the raw gradient onto the spectral subspace that carries its dominant energy, weighted by the singular values $\Sigma_{1:r_t}=\operatorname{diag}(\sigma_1,\dots,\sigma_{r_t})$, and discards the low-energy directions. Retaining $\Sigma$ preserves the energy-weighting of the leading directions, so the reconstruction is a faithful truncation of the gradient rather than an unweighted projection.
The reconstructed gradient then drives an adaptive update in the manner of the AdamW rule of Sec.~\ref{sec:preliminary}: Ares maintains bias-corrected first- and second-moment estimates of $\tilde g_t$, forms the per-coordinate adaptive step, and applies it to the parameter together with decoupled weight decay,
\begin{equation}
\small
    \theta_t = (1 - \eta\lambda)\, \theta_{t-1}
    - \eta\, s \, \frac{\hat m_t}{\sqrt{\hat v_t} + \epsilon},
\end{equation}
where $\eta$ is the learning rate, $\lambda$ the weight-decay coefficient, and $s$ a scale factor for the low-rank update. 
Operating on $\tilde g_t$ confines the update to informative spectral directions, avoiding wasted capacity on the low-energy tail.

This spectral treatment applies only to parameters with a two-dimensional matrix structure, for which a singular spectrum is defined. Ares thus routes by shape: two-dimensional weight matrices receive the low-rank spectral update above, while one-dimensional parameters take a standard AdamW step. The full procedure is summarized in Appendix~\ref{app:pseudocode}.

\subsection{Summarization and Discussion}
\label{sec:discussion}

We summarize Ares in Algorithm~\ref{alg:ares}.
At each step, Ares factorizes gradients with a randomized block power iteration, converts the resulting spectral energy distribution into an effective rank through its entropy, smoothly tracks this rank across steps, and forms an adaptive update confined to the corresponding low-rank subspace, while one-dimensional parameters retain a standard AdamW step.
Its central contribution is that the update rank is neither fixed nor scheduled but \emph{inferred from the observed gradient spectrum} and tracked as it evolves, so the update dimensionality follows the energy concentration, at a per-step cost governed by the target rank rather than the ambient dimension.

\begin{table*}[t]
    \centering
    \small
    \caption{Comparison of Ares with representative optimizers used in large-model training. We distinguish three properties: whether the method restricts the update to a \emph{low-rank} subspace, \emph{how} the rank/update subspace is determined, and whether that determination is \emph{spectral-driven}, \ie inferred from the gradient's spectral energy distribution during training. Ares infers the \emph{dimensionality} of the update subspace from the evolving gradient spectrum.}
    \begin{tabular}{lcccc}
        \toprule
        \textbf{Optimizer} & \textbf{Update rule} & \textbf{Low-rank} & \textbf{Rank} & \textbf{Spectral-driven} \\
        \midrule
        SGD                          & First-order momentum          & No  & ---                 & No \\
        RMSprop                      & Diagonal adaptive             & No  & ---                 & No \\
        Adam~\citep{kingma2015adam}  & Diagonal adaptive             & No  & ---                 & No \\
        AdamW~\citep{loshchilov2019decoupled} & Diagonal adaptive    & No  & ---                 & No \\
        Adafactor~\citep{shazeer2018adafactor} & Factored second moment & No & ---               & No \\
        Lion~\citep{chen2023lion}    & Sign momentum                 & No  & ---                 & No \\
        Sophia~\citep{liu2024sophia} & Diagonal Hessian precond.     & No  & ---                 & No \\
        Shampoo~\citep{gupta2018shampoo} & Full-matrix precond.      & No  & ---                 & No \\
        SOAP~\citep{vyas2024soap}    & Shampoo in Adam eigenbasis    & No  & ---                 & No \\
        Muon~\citep{jordan2024muon}  & Orthogonalized momentum       & No  & ---                 & Direction only \\
        \midrule
        GaLore~\citep{zhao2024galore} & Low-rank projection          & Yes & Constant $r$ & No \\
        Rank schedule                & Low-rank projection           & Yes & Preset $r(t)$        & No \\
        \textbf{Ares} (ours)         & Low-rank, spectral-driven     & Yes & Spectrum & Yes \\
        \bottomrule
    \end{tabular}
    \label{tab:rank_taxonomy}
\end{table*}

Table~\ref{tab:rank_taxonomy} places Ares among representative optimizers along three axes: whether the update is restricted to a low-rank subspace, how the rank is determined, and whether that determination is spectral-driven.
First-order, diagonal adaptive, and preconditioned methods (\eg SGD, AdamW, Sophia, SOAP) all apply full-dimensional updates, adapting only the step size or a preconditioner, and Muon adapts the update \emph{direction} but leaves its dimensionality unconstrained.
Among low-rank methods, GaLore fixes the rank to a constant and a rank schedule varies it along a preset curve, so in both cases the rank is chosen independently of the observed gradient.
Ares is thus the one method in Table~\ref{tab:rank_taxonomy} that is low-rank \emph{and} spectral-driven, with $r_t$ read online from the spectrum and tracked as it evolves. Concurrent work has begun to study low-rank and spectral structure specifically in RLVR~\citep{NExt,Fan2026Rethinking}; relative to these, our distinguishing property is that the rank is neither fixed nor scheduled but inferred from the gradient's spectral energy. This is precisely the property motivated in Sec.~\ref{sec:intro} and constitutes the core contribution of this work: matching the optimizer to the evolving gradient geometry rather than to a fixed rank.

\section{Experiment}

\subsection{Experiment Settings}
\label{sec:exp_setup}

\paratitle{Dataset.}
We conduct RLVR on the DAPO dataset~\citep{yu2025dapo}, a large-scale collection of competition-level mathematics problems paired with verifiable answers. 
To assess generalization, we evaluate on seven widely used mathematical reasoning benchmarks that span a range of difficulty: AIME24~\citep{aime24}, AIME25~\citep{aime25}, AIME26~\citep{aime26}, AMC23~\citep{amc23}, MATH500~\citep{math500}, Minerva Math~\citep{minerva}, and OlympiadBench (OlyBench)~\citep{olympiadbench}.
Following common practice, we report the average accuracy over $8$ sampled responses per problem to reduce evaluation variance.

\paratitle{Baselines.}
We compare Ares against a representative set of optimizers commonly used for
LLM training. 
These include (1) standard first-order and adaptive optimizers, \eg SGD, RMSprop, Adam~\citep{kingma2015adam}, and AdamW~\citep{loshchilov2019decoupled}; and (2) a low-rank optimizer, GaLore~\citep{zhao2024galore}, which projects gradients onto a subspace of externally fixed rank, and Muon~\citep{jordan2024muon}, which orthogonalizes the momentum update. 
All baselines are trained under an identical RLVR pipeline.
We privode the implementation details in Appendix~\ref{app:implementation}.

\subsection{Main Results}

\begin{table*}[t]
    \centering
    \small
    \caption{Accuracy of LLMs trained via GRPO with various optimizers. ``Avg.'' refers to average accuracy. The best is in bold and the second is underlined.}
      \begin{tabular}{lcccccccc}
        \toprule
        \textbf{Optimizer} & \textbf{AIME24} & \textbf{AIME25} & \textbf{AIME26} & \textbf{AMC23} & \textbf{MATH500} & \textbf{Minerva} & \textbf{OlyBench} & \textbf{Avg.} \\
        \midrule
        \multicolumn{9}{c}{\textit{Qwen3-1.7B-Base}} \\
        AdamW & 7.9 & 4.2 & 5.0 & 37.8 & 59.2 & 15.5 & 25.7 & 22.2 \\
        SGD & 0.4 & 0.0 & 0.4 & 2.8 & 6.3 & 2.9 & 1.3 & 2.0 \\
        RMSprop & 0.0 & 0.0 & 0.0 & 13.1 & 33.5 & 4.8 & 9.6 & 8.7  \\
        Adam & 7.5 & \underline{6.3} & 4.6 & \underline{40.0} & \underline{59.3} & 15.3 & 25.1 & 22.6  \\
        GaLore & \underline{8.3} & \underline{6.3} & \underline{6.3} & 37.8 & 58.9 & \underline{16.9} & \underline{26.1} & \underline{22.9}  \\
        Muon & 0.4 & 0.4 & 0.0 & 3.4 & 7.8 & 2.8 & 1.4 & 2.3  \\
        Ares (Ours) & \textbf{15.0} & \textbf{10.8} & \textbf{10.0} & \textbf{43.4} & \textbf{61.6} & \textbf{20.4} & \textbf{29.4} & \textbf{27.2}  \\
        \midrule
        \multicolumn{9}{c}{\textit{Qwen3-4B-Base}} \\
        AdamW & \underline{21.3} & 16.3 & 13.8 & \underline{60.9} & \underline{71.9} & \underline{27.4} & \underline{39.1} & \underline{35.8} \\
        SGD & 0.8 & 0.4 & 0.4 & 6.6 & 14.5 & 5.7 & 3.8 & 4.6 \\
        RMSprop & 2.5 & 0.8 & 2.5 & 13.1 & 38.7 & 7.6 & 12.9 & 11.2 \\
        Adam & 18.3 & \underline{16.7} & \underline{15.4} & 57.2 & 70.6 & 23.3 & 36.8 & 34.0 \\
        GaLore & 15.8 & 13.3 & 12.9 & 55.9 & 70.8 & 23.4 & 37.2 & 32.8 \\
        Muon & 1.3 & 0.4 & 0.4 & 8.8 & 17.5 & 6.6 & 5.4 & 5.8 \\
        Ares (Ours) & \textbf{25.4} & \textbf{17.5} & \textbf{20.0} & \textbf{65.0} & \textbf{73.4} & \textbf{28.2} & \textbf{40.3} & \textbf{38.5} \\
        \midrule
        \multicolumn{9}{c}{\textit{Qwen3-8B-Base}} \\
        AdamW & \underline{25.8} & 19.6 & 19.2 & \underline{68.4} & \textbf{75.0} & \underline{27.7} & \underline{42.6} & \underline{39.8} \\
        SGD & 0.8 & 1.3 & 0.8 & 10.9 & 21.8 & 8.0 & 7.1 & 7.2 \\
        RMSprop & 2.0 & 3.3 & 1.3 & 18.8 & 24.7 & 10.7 & 12.2 & 10.4 \\
        Adam & 23.3 & 19.6 & 17.5 & 67.8 & 74.1 & 22.7 & 40.9 & 38.0 \\
        GaLore & 25.4 & \underline{20.8} & \underline{20.4} & 66.9 & 73.2 & 24.5 & 40.3 & 38.8 \\
        Muon & 2.1 & 3.3 & 2.5 & 15.6 & 23.2 & 7.1 & 8.5 & 8.9 \\
        Ares (Ours) & \textbf{28.8} & \textbf{22.5} & \textbf{21.7} & \textbf{73.8} & \underline{74.9} & \textbf{29.4} & \textbf{43.4} & \textbf{42.1} \\
        \bottomrule
    \end{tabular}
    \label{tab:main_results}
\end{table*}

Table~\ref{tab:main_results} reports the performance of different optimizers with GRPO. We summarize four findings.

First, Ares delivers consistent and substantial gains across model scales. Ares attains the best average accuracy at every scale, \ie $27.2$, $38.5$, and $42.1$ on the 1.7B, 4B, and 8B models, and ranks first on the large majority of the seven individual benchmarks. It improves over the widely used AdamW by $5.0$, $2.7$, and $2.3$ points. The improvement is therefore not confined to a particular setting but holds broadly across benchmarks and model sizes.

Second, the advantage is stable as the model scales up. Rather than shrinking with capacity, the lead of Ares persists from 1.7B to 8B, and it remains the top method at every scale. This indicates that the benefit of matching the optimizer to the RLVR gradient geometry does not vanish as the base model grows stronger, so the approach scales gracefully rather than being an artifact of small models.

Third, naively exploiting low-rank structure is insufficient to capture the gradient characteristics of RLVR. GaLore, which projects gradients onto a subspace of externally fixed rank, is merely on par with AdamW and is in fact worse at the 4B scale ($32.8$ vs. $35.8$). Imposing a low-rank constraint alone is thus not enough: because the spectral concentration of RLVR gradients evolves during training, the update rank must be inferred from the gradient spectrum and adapted online rather than fixed in advance, which is precisely what distinguishes Ares from GaLore.

Finally, the degree to which the optimizer matches RLVR dynamics is critical to final performance. 
The optimizers differ markedly from one another: SGD, RMSprop, and Muon are consistently weak across all three scales, collapsing to near-zero or single-digit accuracy, whereas AdamW and GaLore are more stable. 
Against this backdrop, Ares improves further at every scale. 
Taken together, these gaps show that the optimizer is not an interchangeable building block in RLVR; its design must be tailored to the distinctive and dynamically evolving gradient geometry that RLVR induces.

\subsection{Detailed Analysis}

\begin{table*}[t]
    \centering
    \small
    \setlength{\tabcolsep}{3.5pt}
    \caption{Ablation study of Ares on Qwen3-1.7B-Base on mathematical tasks. ``Avg.'' refers to average accuracy. Further results of various models can be found in Table~\ref{tab:ablation_study_full}.}
      \begin{tabular}{lcccccccc}
        \toprule
        \textbf{Optimizer} & \textbf{AIME24} & \textbf{AIME25} & \textbf{AIME26} & \textbf{AMC23} & \textbf{MATH500} & \textbf{Minerva} & \textbf{OlyBench} & \textbf{Avg.} \\
        \midrule
        Ares (Ours) & 15.0  & 10.8  & 10.0  & 43.4  & 61.6  & 20.4  & 29.4  & 27.2 \\
        \quad \textit{w/o} Rank Momentum & 10.8  & 7.5  & 5.8  & 41.9  & 57.9  & 18.2  & 26.7  & 24.1 \\
        \quad \textit{w/o} Dynamic Rank & 12.9  & 9.2  & 5.0  & 48.1  & 61.9  & 19.7  & 29.5  & 26.6 \\
        \quad \textit{w/o} Low Rank & 0.0  & 0.0  & 0.4  & 4.4  & 9.1  & 2.9  & 2.0  & 2.7 \\
        \quad \textit{w/o} Spectral Normalize & 10.8  & 6.7  & 5.0  & 48.1  & 61.9  & 20.0  & 28.0  & 25.8 \\
        \quad \textit{w/o} Adaptive LR & 10.0  & 5.0  & 4.2  & 39.4  & 60.5  & 17.4  & 25.9  & 23.2 \\
        \bottomrule
    \end{tabular}
    \label{tab:ablation_study}
\end{table*}

\paratitle{Ablation Study.}
Table~\ref{tab:ablation_study} isolates each component of Ares. 
Removing any single component lowers accuracy, so the modules are complementary.
The low-rank update is the core: \textit{w/o Low Rank} collapses accuracy to $2.7$ and this collapse persists even with the other mechanisms active, as adjusting the learning rate or fixing $\sigma\!=\!1$ (an orthogonalized step in the spirit of Muon) brings no benefit without it; confining the update to the low-rank spectral subspace is thus the foundation that makes every other component useful. 
Dynamic rank and momentum, in turn, act as one mechanism: \textit{w/o Dynamic Rank} ($26.6$) alone appears to gain little, because a rank estimated from noisy on-policy gradients fluctuates and destabilizes the update subspace, but momentum smooths this estimate and stabilizes the subspace, so that \textit{w/o Momentum} drops more sharply to $24.1$.
Their combination turns a fluctuating rank into a reliable gain, showing that the pieces reinforce one another rather than forming a loose collection of independently useful tricks.

\begin{figure*}[t]
    \includegraphics[width=\linewidth]{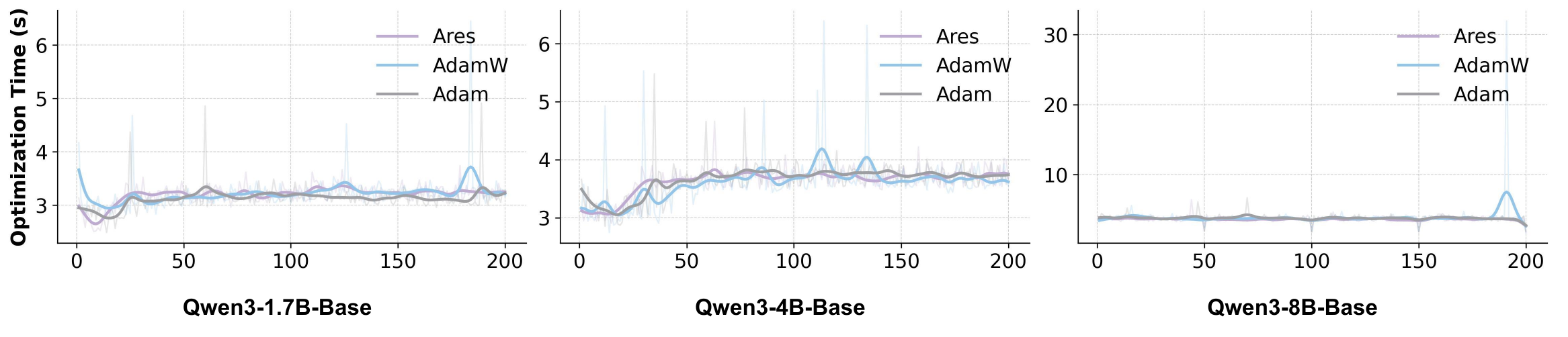}
    \caption{The time to update model parameters expended by various optimizers.}
    \label{fig:computational_consumption}
\end{figure*}

\paratitle{Analysis of Computational Consumption.}
Fig.~\ref{fig:computational_consumption} reports the per-step time each optimizer spends updating the parameters.
Although Ares performs extra computation over a plain adaptive step, estimating the spectral energy and forming the low-rank subspace, this is kept cheap by using randomized subspace iteration~\citep{halko2011finding} to extract only the leading $r$ components rather than a full SVD. As a result, its update time stays close to standard optimizers and far below an exact spectral method, confirming that the spectral-driven design adds only low-order overhead.
The low-rank factorization also opens a path to lower resource usage, since the first- and second-moment states could in principle be maintained in a factored form~\citep{zhao2024galore}. We stress, however, that our current implementation does not yet realize this saving, which maintains dense moment tensors as in AdamW, so the present contribution is update quality rather than optimizer-state memory, which we leave for future work.

\begin{figure}[t]
    \includegraphics[width=\linewidth]{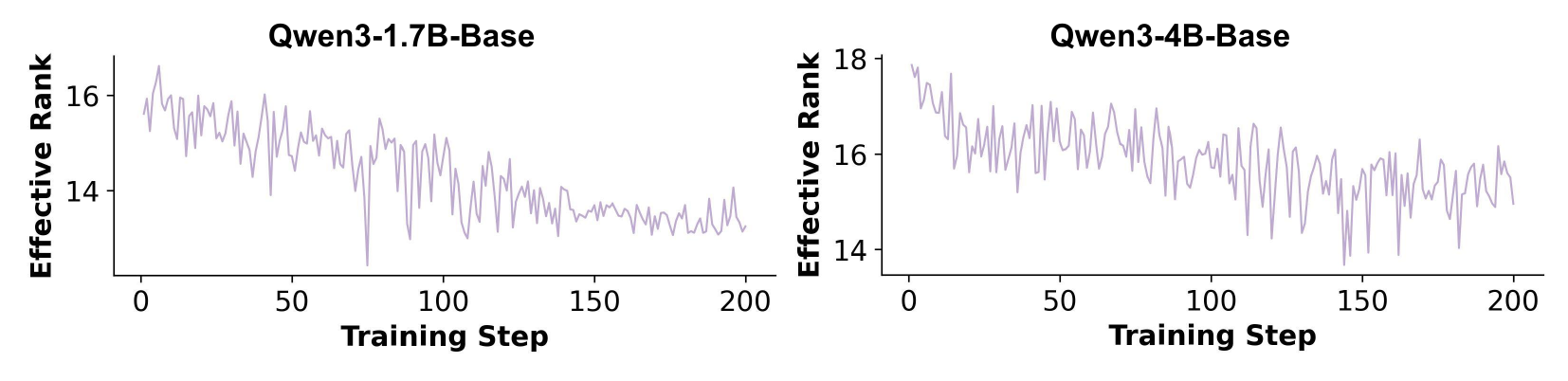}
    \caption{The dynamics of the effective rank of the gradient during the RLVR procedure.}
    \label{fig:energy_dynamic}
\end{figure}

\paratitle{Dynamics of Energy.}
Fig.~\ref{fig:energy_dynamic} tracks the effective rank of the gradient, estimated from its spectral energy distribution over the RLVR. 
First, the effective rank declines steadily as training proceeds. 
The gradient energy thus concentrates into progressively fewer directions during RLVR, and the effective dimensionality of the useful update shrinks over time. 
This directly corroborates the low-rank characterization of Sec.~\ref{sec:intro} on the very quantity Ares acts on, and it is what justifies inferring the update rank from the spectrum and letting it decay online rather than fixing it in advance. 
Second, the larger model operates at a consistently higher effective rank: the 4B curve sits above the 1.7B curve throughout training, both at initialization and after convergence. 
The larger model spreads its gradient energy over more effective directions, suggesting that greater capacity lets it explore and exploit a richer set of useful update directions and thereby encode more knowledge. 
This also carries a practical implication for Ares: because the appropriate rank grows with model capacity, inferring it from the observed spectrum, is what allows the same optimizer to adapt across model scales.

\begin{wrapfigure}{r}{0.4\linewidth}
    \includegraphics[width=\linewidth]{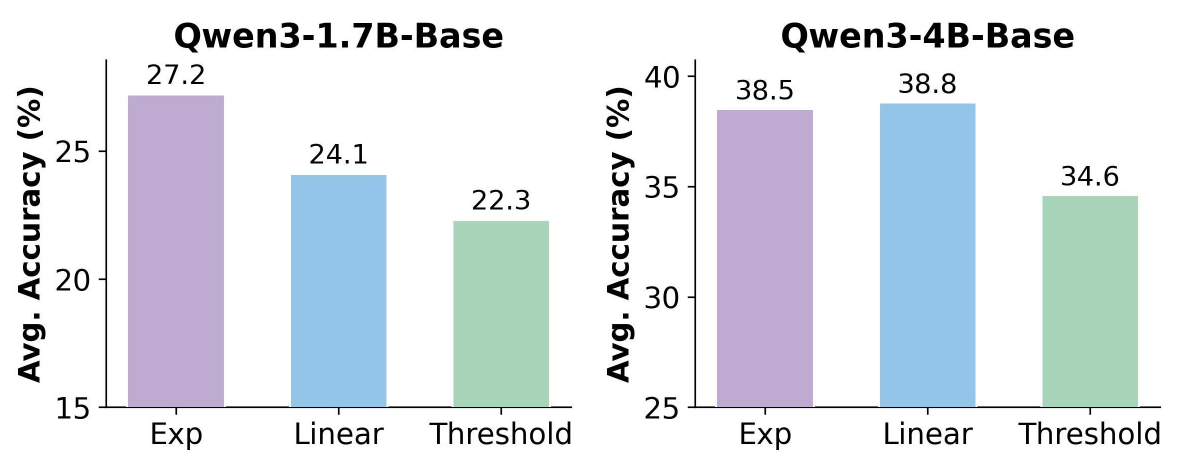}
    \caption{Accuracy of various adaptation methods. Full results in Table~\ref{tab:rank_adaptation_full}}
    \label{fig:rank_adaptation}
\end{wrapfigure}

\paratitle{Effect of Different Rank Adaptation Strategies.}
Given the estimated spectral energy, Ares still needs a rule to translate it into a concrete rank. We compare three strategies in Fig.~\ref{fig:rank_adaptation}: \textit{Exp} maps energy to rank exponentially, \textit{Linear} in direct proportion, and \textit{Threshold} sets the rank at the smallest value whose cumulative energy reaches a fixed fraction, an abrupt cutoff.
The exponential mapping used in Ares performs best overall, attaining the top or near-top accuracy at both scales and never being clearly beaten, which makes it a robust default.
Smaller models are markedly more sensitive to the choice: the three strategies span a wide range on Qwen3-1.7B but are nearly indistinguishable among the smooth mappings on Qwen3-4B, as a smaller model has little margin to absorb a poorly chosen update subspace.
\textit{Threshold} is consistently worst, since fixed energy thresholds over-admit weak directions under long-tailed spectra, whereas a smooth mapping adapts rank continuously to spectral shape, motivating Ares's exponential form.
\section{Conclusion}

In this work, we revisited optimizer design for RLVR and show that its gradients exhibit a distinctive, evolving spectral geometry: energy concentrates progressively in leading singular components, driving the effective update rank downward, in contrast to the SFT and random-reward controls we study.
Based on this observation, we introduced \textbf{Ares}, which automatically infers and smoothly adapts its update rank from the gradient spectrum while confining the gradient signal of each update to the resulting low-rank subspace.
Extensive experiments across model scales, tasks, and RLVR algorithms demonstrated consistent gains over widely used optimizers, with ablations supporting the contribution of each component.
Our findings pointed to a broader principle: effective optimizers should adapt to the gradient geometry induced by the training paradigm, rather than inherit assumptions from supervised training.


\bibliography{ref}

\newpage

\appendix

\section{Implementation Details}
\label{app:implementation}

We build on the Qwen3-Base models~\citep{yang2025qwen3} at three scales, \ie 1.7B, 4B, and 8B, and train all optimizers with GRPO~\citep{shao2024deepseekmath} under the same configuration.
For each training step, we sample a group of $16$ responses per prompt with a rollout batch size of $128$ and a maximum generation length of $8192$ tokens, and update the policy with a mini-batch size of $32$.
We tune the learning rate for each optimizer separately, searching over $\{1,2,5\}\times 10^{-6}$ and $\{1,2\}\times 10^{-5}$ with the same budget for every optimizer (AdamW, GaLore, Muon, and Ares) and selecting by final training accuracy, and use $5\times 10^{-6}$ for Ares.
For the low-rank optimizers, the rank is set to $64$; for Ares, we initialize the rank at $64$ and adapt it online from the gradient spectrum.
For the Ares-specific hyperparameters we use RBPI oversampling $p=10$ and power iterations $q=20$ (Appendix~\ref{app:rbpi}), rank-EMA smoothing $c=0.9$, and scale $s=1$, shared across all layers and model scales rather than tuned per layer; the rank-to-index mapping uses nearest-integer rounding.
All models are trained for $300$ steps.

\section{Pseudo Code of Ares}
\label{app:pseudocode}

Algorithm~\ref{alg:ares} summarizes one optimizer step of Ares, following the notation of Sec.~\ref{sec:approach}. Two-dimensional weights receive the spectral low-rank update (Sections~\ref{sec:factorization}--\ref{sec:lowrank_opt}), while one-dimensional parameters fall back to a standard AdamW step. State maintained across steps consists of the moment estimates $(m, v)$ and the tracked rank $r$; the momentum coefficients $(\beta_1,\beta_2)$ follow the AdamW convention of Sec.~\ref{sec:preliminary}.

\begin{algorithm}[H]
\small
\DontPrintSemicolon
\SetKwInput{KwState}{State}
\caption{Ares: one optimizer step for a single parameter $\theta$}
\label{alg:ares}
\KwIn{gradient $g_t$; step $t$; learning rate $\eta$, weight decay $\lambda$, scale $s$, epsilon $\epsilon$; smoothing $c$; target rank $k$; RBPI parameters $(p,q)$}
\KwState{first/second moments $m,v$; uncorrected rank EMA accumulator $z$ (init.\ $z\leftarrow 0$); tracked rank $r$ (init.\ $r\leftarrow k$)}
\eIf{$\theta$ is a $2$D weight matrix}{
  $U,\Sigma,V \leftarrow \textsc{RBPI}(g_t,\,k,\,p,\,q)$ \tcp*{leading singular subspace (Sec.~\ref{sec:factorization})}
  $p_i \leftarrow \sigma_i^2 / \sum_j \sigma_j^2$;\quad $H \leftarrow -\sum_i p_i \log p_i$;\quad $\hat r_t \leftarrow \exp(H)$ \tcp*{energy entropy (Sec.~\ref{sec:rank_estimation})}
  $z \leftarrow c\,z + (1-c)\,\hat r_t$;\quad $r \leftarrow \operatorname{clip}\!\big(\operatorname{round}(z / (1-c^{\,t})),\, 1,\, k\big)$ \tcp*{smooth and track}
  $\tilde g_t \leftarrow U_{:,\,1:r}\, \Sigma_{1:r}\, V^\top_{1:r,\,:}$ \tcp*{rank-$r$ truncated-SVD gradient}
}{
  $\tilde g_t \leftarrow g_t$ \tcp*{$1$D parameter: standard AdamW}
}
$m \leftarrow \beta_1 m + (1-\beta_1)\tilde g_t$;\quad $v \leftarrow \beta_2 v + (1-\beta_2)\tilde g_t^2$ \tcp*{moment updates}
$\hat m \leftarrow m/(1-\beta_1^t)$;\quad $\hat v \leftarrow v/(1-\beta_2^t)$ \tcp*{bias correction}
$s_\theta \leftarrow s$ if $\theta$ is $2$D else $1$ \tcp*{low-rank scaling only for matrices}
$\theta \leftarrow (1-\eta\lambda)\,\theta - \eta\, s_\theta\, \hat m/(\sqrt{\hat v}+\epsilon)$ \tcp*{decoupled weight decay + update}
\Return{$\theta$}
\end{algorithm}

\section{Analysis of the Randomized Block Power Iteration}
\label{app:rbpi}

At the core of Ares is a randomized block power iteration (RBPI) that estimates, at every optimizer step, the leading singular subspace of the (momentum-smoothed) gradient matrix $A\in\mathbb{R}^{m\times n}$. Rather than computing a full singular value decomposition (SVD), Ares uses RBPI as an approximation whose leading-order cost is governed by the target rank $k$ rather than by the smaller ambient dimension $\min(m,n)$. This appendix analyzes (i) the computational cost of RBPI relative to a dense SVD, and (ii) the sense in which RBPI approximates the leading spectral subspace, together with the conditions under which the approximation is expected to be accurate.

Throughout, we denote the input matrix by $A\in\mathbb{R}^{m\times n}$, and without loss of generality we assume $m\ge n$ (the implementation transposes internally, so the analysis is symmetric in $m$ and $n$). We write $r=\min(m,n)$ for the full rank, $k$ for the target rank, $p$ for the oversampling parameter (\texttt{oversample}, default $p=10$), and $q$ for the number of power iterations (\texttt{n\_iter}, default $q=20$). We set $\ell=k+p$ for the size of the sampled block. The singular values of $A$ are $\sigma_1\ge\sigma_2\ge\cdots\ge\sigma_r\ge 0$, and by the Eckart--Young theorem the best rank-$k$ approximation error is $\|A-A_k\|_2=\sigma_{k+1}$ and $\|A-A_k\|_F^2=\sum_{j>k}\sigma_j^2$, where $A_k$ is the truncation of the SVD to its top $k$ components.

Before analyzing the two aspects above, we note that the routine is not a one-pass randomized SVD but rather Halko et al.'s \emph{power scheme}~\citep{halko2011finding}: after the initial sketch $Y=A\Omega$, the loop alternately applies $A^\top$ and $A$ with a QR re-orthonormalization after each application, repeated $q$ times, so that the final block satisfies $Y_q\approx (AA^\top)^q A\,\Omega$. Since $(AA^\top)^q A$ has the same singular vectors as $A$ but singular values $\sigma_i^{2q+1}$, the scheme amplifies the separation between the leading and trailing spectrum, which is the property the approximation analysis below relies on.

\subsection{Time Complexity}
\label{app:rbpi_complexity}

\paragraph{Dense SVD.}
Computing the thin SVD $A=U\Sigma V^\top$ of a dense $m\times n$ matrix with $m\ge n$ costs
\begin{equation}
    T_{\mathrm{SVD}} = \Theta\!\left(m n^2\right)
\end{equation}
for the standard bidiagonalization-based pipeline. This is the cost model of the dense driver behind \texttt{torch.linalg.svd}; the key point for our purposes is that, for such a routine, the cost scales with the full smaller dimension $n$ rather than with the target rank $k$, so requesting only the top $k\ll n$ components does not by itself reduce the work. (Some backends offer specialized truncated or approximate SVD drivers; we do not rely on their availability here.)

\paragraph{Randomized block power iteration.}
We account for RBPI step by step, following the implementation. Matrix--matrix products dominate; QR of an $m\times\ell$ (respectively $n\times\ell$) matrix costs $O(m\ell^2)$ (respectively $O(n\ell^2)$), and the small dense SVD of the $\ell\times n$ matrix $B$ costs $O(n\ell^2)$.
\begin{itemize}
    \item \textbf{Sketch} $Y=A\Omega$ with $\Omega\in\mathbb{R}^{n\times\ell}$: $O(mn\ell)$.
    \item \textbf{Power iterations} ($q$ rounds): each round forms $A^\top Y$ and $AY$ at $O(mn\ell)$ each, plus two QR factorizations at $O((m+n)\ell^2)$; total $O\!\left(q\,(mn\ell+(m+n)\ell^2)\right)$.
    \item \textbf{Orthonormalization} $Q$ of $Y$: $O(m\ell^2)$.
    \item \textbf{Projection} $B=Q^\top A\in\mathbb{R}^{\ell\times n}$: $O(mn\ell)$.
    \item \textbf{Small SVD} of $B$: $O(n\ell^2)$.
    \item \textbf{Lift} $U=Q\,U_B$: $O(m\ell^2)$.
\end{itemize}
Collecting the terms, the total cost is
\begin{equation}
    T_{\mathrm{RBPI}} = O\!\left((q+1)\,mn\ell + (q+1)(m+n)\ell^2\right).
\end{equation}
When $\ell\le n\le m$ we have $(m+n)\ell^2\le 2m\ell^2\le 2mn\ell$, so the $mn\ell$ terms dominate and the bound simplifies to
\begin{equation}
    T_{\mathrm{RBPI}} = O\!\left(q\,mn\ell\right),
    \qquad \ell=k+p .
\end{equation}

\paragraph{Rank-dependent cost and the resulting ratio.}
Treating the two costs as $T_{\mathrm{SVD}}=\Theta(mn^2)$ and $T_{\mathrm{RBPI}}=\Theta(q\,mn\ell)$ under the assumption $\ell\le n$ (so that the leading terms above are indeed dominant), the leading-order work ratio is
\begin{equation}
    \frac{T_{\mathrm{SVD}}}{T_{\mathrm{RBPI}}}
    = \Theta\!\left(\frac{n}{q\,(k+p)}\right).
\end{equation}
The essential distinction is that the dense SVD grows quadratically in the smaller ambient dimension $n$, whereas the dominant matrix products in RBPI grow only linearly in the block size $\ell=k+p$. In other words, RBPI trades a dependence on $n^2$ for a dependence on $q(k+p)\,n$, and the asymptotic advantage becomes substantial precisely when $q(k+p)\ll n$. This is the regime Ares targets: gradient matrices of transformer weight layers have $n$ on the order of $10^3$--$10^4$, while the target rank is a few tens (initial rank $64$ in our experiments), so $k+p$ is typically much smaller than $n$. We stress that $q$ is a fixed constant with respect to the matrix size rather than a negligible one, with $q=20$ each power iteration performs two matrix products, so the constant is not small, and the leading-order ratio $n/[q(k+p)]$ should be read with this factor in mind rather than as an unconditional order-of-magnitude gain.

\paragraph{Practical remark.}
RBPI is dominated by GEMM-type operations (the sketch, the projection, and the two products per power iteration), which are highly optimized on modern GPUs. This is a favorable property, but it does not by itself imply a wall-clock speedup: RBPI also incurs repeated QR factorizations, normalizations, and kernel-launch and synchronization overhead, while dense SVD kernels are themselves highly tuned. The realized wall-clock speedup therefore depends on the matrix shape, the hardware, the implementation, and the number of power iterations, and is best assessed empirically rather than inferred from the flop count alone.

\subsection{Approximation Guarantee}
\label{app:rbpi_correctness}

We now discuss the sense in which RBPI approximates the leading singular subspace of $A$. The routine is a standard instance of the randomized subspace-iteration framework of \citet{halko2011finding}, and its behavior rests on two mechanisms: the range of the sketch tends to align with the dominant singular directions, and each power iteration sharpens this alignment.

\paragraph{Exactness in the noiseless low-rank case.}
First consider the ideal case in which $A$ has exact rank $k$, \ie $\sigma_{k+1}=\cdots=\sigma_r=0$. Then $A=U_k\Sigma_k V_k^\top$ with $U_k\in\mathbb{R}^{m\times k}$, and the sketch $Y=A\Omega=U_k(\Sigma_k V_k^\top\Omega)$ lies in $\operatorname{range}(U_k)$. As long as $V_k^\top\Omega\in\mathbb{R}^{k\times\ell}$ has full row rank $k$, which holds almost surely for a Gaussian $\Omega$ once $\ell\ge k$, the columns of $Y$ span $\operatorname{range}(U_k)$ exactly. Consequently, the orthonormal factor $Q$ satisfies $QQ^\top A=A$, so $B=Q^\top A$ and its SVD reproduces the true singular triplets, and the lifted factor $U=Q\,U_B$ equals $U_k$ up to sign/rotation. In this regime RBPI is exact, using only $\ell=k+p$ sketch columns rather than operating on all $n$ columns.

\paragraph{Error bound in the general case.}
When $A$ is only approximately low rank, the projection error is controlled by the trailing singular values. Let $Q$ be the orthonormal basis produced after $q$ power iterations. Corollary~10.10 of \citet{halko2011finding}, combined with the bound $\sigma_j\le\sigma_{k+1}$ for $j>k$, gives, in expectation over the Gaussian test matrix $\Omega$ with oversampling $p\ge 2$,
\begin{equation}
    \label{eq:rbpi_bound}
    \mathbb{E}\,\big\|A-QQ^\top A\big\|_2
    \;\le\;
    \left[\,1+\sqrt{\tfrac{k}{p-1}}
    + \tfrac{e\sqrt{k+p}}{p}\sqrt{\,r-k\,}\,\right]^{\!\frac{1}{2q+1}}
    \sigma_{k+1}.
\end{equation}
Two features of this bound are worth noting. \emph{(i) Relation to the optimum:} the right-hand side is a multiplicative factor times $\sigma_{k+1}$, which by Eckart--Young is exactly the error of the best rank-$k$ approximation, so the projection error is bounded relative to this optimal rank-$k$ spectral error. (The projection $QQ^\top A$ itself has rank at most $\ell=k+p$, so this is a comparison of the rank-$\ell$ projection error against the best rank-$k$ error, not a claim that RBPI returns the optimal rank-$k$ approximation.) \emph{(ii) Effect of $q$:} the bracketed factor is raised to the power $1/(2q+1)$, so as $q$ increases this factor decreases toward $1$ and the bound approaches $\sigma_{k+1}$, reflecting the increasingly sharp spectral filtering induced by the power scheme. The rate at which the trailing spectrum is suppressed is governed by the spectral separation: because the power scheme maps $\sigma_j\mapsto\sigma_j^{2q+1}$, the relative contribution of trailing components is attenuated at a rate governed by powers of $\sigma_{k+1}/\sigma_k$, so a larger gap yields faster suppression. This is the setting Ares operates in: RLVR gradients are empirically spectrally concentrated (Sec.~\ref{sec:intro}), so the gap below the leading components is typically large and a moderate $q$ can drive the bracketed factor close to $1$.

\paragraph{Role of oversampling.}
The oversampling parameter $p$ enters \eqref{eq:rbpi_bound} through the terms $\sqrt{k/(p-1)}$ and $\tfrac{e\sqrt{k+p}}{p}\sqrt{r-k}$, both of which decrease as $p$ increases. Intuitively, drawing $\ell=k+p>k$ random directions makes the sampled block $V_k^\top\Omega$ better conditioned and more likely to capture all $k$ leading directions, providing robustness against a near-degenerate random projection. A small constant oversample, such as the $p=10$ used in our implementation, adds robustness to the randomized projection while increasing the cost only linearly in $\ell=k+p$, which is absorbed into the $O(q\,mn\ell)$ term above. We state this as an expected-error effect and do not claim a specific failure probability, which would require a separate high-probability analysis.

\paragraph{From subspace approximation to low-rank reconstruction.}
The projection bound \eqref{eq:rbpi_bound} translates directly into a reconstruction bound for the compressed representation. Let $B=Q^\top A$ and compute its exact SVD $B=U_B\Sigma V^\top$. Then
\begin{equation}
    Q B = Q\,U_B\,\Sigma V^\top = QQ^\top A,
\end{equation}
and, because $Q$ has orthonormal columns and hence preserves the spectral norm of the residual,
\begin{equation}
    \big\|A - Q\,U_B\,\Sigma V^\top\big\|_2 = \big\|A - QQ^\top A\big\|_2 .
\end{equation}
Thus the randomized subspace bound directly controls the reconstruction error of the compressed representation used by Ares. This is a statement about the reconstruction $Q U_B \Sigma V^\top$ as a whole; it does not by itself imply that each returned singular vector or singular value matches its true counterpart with the same error, since the accuracy of individual singular vectors additionally depends on the spectral gap through perturbation results of the Davis--Kahan/Wedin type. When the top-$k$ singular values are well separated from the remainder, \ie the regime RLVR gradients empirically exhibit, such perturbation results further imply accurate recovery of the associated singular vectors and values, which are the quantities Ares uses to form the low-rank update and to estimate the effective rank from the spectral energy entropy.

\subsection{Implication for Ares}
\label{app:rbpi_summary}

In summary, RBPI reduces the rank dependence of extracting the leading spectral subspace from $n^2$ to $q(k+p)\,n$, \ie from $O(mn^2)$ to $O(q\,mn\ell)$ with $\ell=k+p$. The corresponding leading-order work ratio scales as $n/[q(k+p)]$, which is favorable when $q(k+p)\ll n$, though the constant contributed by $q$ and the practical overheads mean the realized speedup should be verified empirically rather than asserted from the flop count. On the accuracy side, RBPI is exact when $A$ is genuinely rank-$k$, and in the general case its projection error is bounded by a multiplicative factor times the optimal rank-$k$ error $\sigma_{k+1}$, with the factor decreasing toward $1$ as the number of power iterations grows and as the spectral gap widens; oversampling adds robustness to the random projection. Because RLVR gradients empirically display a pronounced spectral gap (Sec.~\ref{sec:intro}), a moderate number of power iterations and a small oversample are expected to let RBPI track the leading subspace closely, which is what makes it a suitable and efficient replacement for a dense SVD inside Ares.

\section{Further Analysis of Ares}

\subsection{Comparison of Training Efficiency}

\begin{figure*}[t]
    \includegraphics[width=\linewidth]{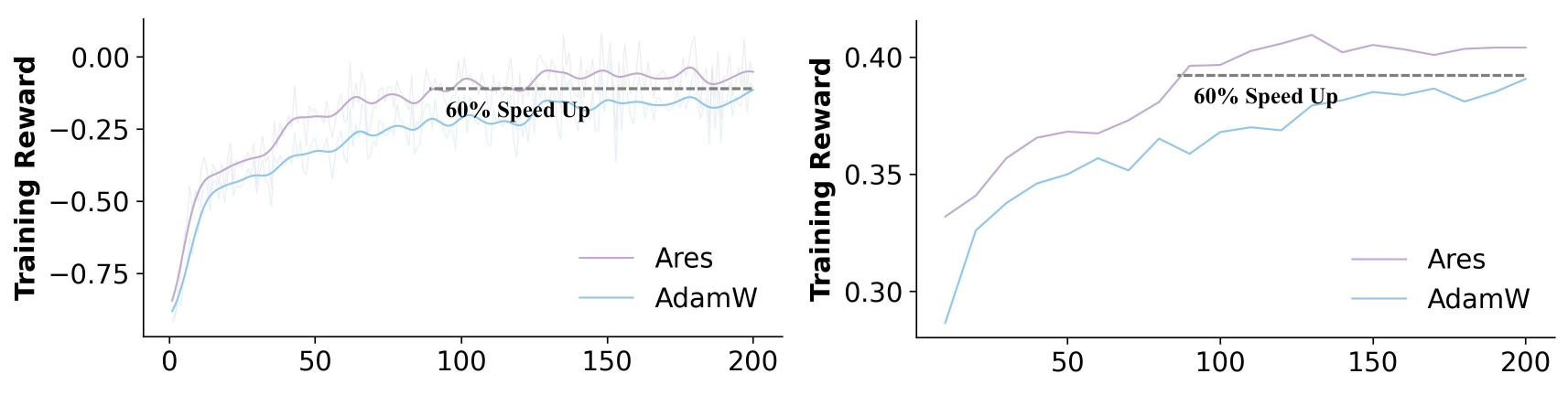}
    \caption{The comparison of training efficiency on Qwen3-4B-Base between ours Ares and AdamW.}
    \label{fig:training_efficiency}
\end{figure*}

Fig.~\ref{fig:training_efficiency} compares the training dynamics of Ares and AdamW on Qwen3-4B-Base. Ares attains the same accuracy as AdamW in substantially fewer training steps and continues to improve after AdamW has plateaued, converging to a higher final accuracy. The comparison here is in terms of optimizer steps; since both methods share the same rollout and reward-verification pipeline per step and the added optimizer overhead is modest (Fig.~\ref{fig:computational_consumption}), the step-level efficiency translates into a comparable wall-clock advantage, though we do not claim a fully compute-matched speedup. This efficiency gain is a direct consequence of the low-rank geometry identified in Sec.~\ref{sec:intro}: because the energy of RLVR gradients concentrates in a few leading singular components, restricting the gradient signal to the spectral subspace that carries this energy suppresses uninformative directions and yields a more effective update per step. Optimizing within the geometry that RLVR gradients actually exhibit therefore translates into faster and better convergence.

\subsection{Detailed Results of Ablation Study}

\begin{table*}[t]
    \centering
    \small
    \setlength{\tabcolsep}{2.5pt}
    \caption{Accuracy of Qwen models trained via GRPO with various optimizers on mathematical tasks. ``Avg.'' refers to average accuracy. The best is in bold and the second is underlined.}
      \begin{tabular}{lcccccccc}
        \toprule
        \textbf{Optimizer} & \textbf{AIME24} & \textbf{AIME25} & \textbf{AIME26} & \textbf{AMC23} & \textbf{MATH500} & \textbf{Minerva} & \textbf{OlyBench} & \textbf{Avg.} \\
        \midrule
        \multicolumn{9}{c}{\textit{Qwen3-1.7B-Base}} \\
        Ares (Ours) & 15.0  & 10.8  & 10.0  & 43.4  & 61.6  & 20.4  & 29.4  & 27.2 \\
        \quad \textit{w/o} Rank Momentum & 10.8  & 7.5  & 5.8  & 41.9  & 57.9  & 18.2  & 26.7  & 24.1 \\
        \quad \textit{w/o} Dynamic Rank & 12.9  & 9.2  & 5.0  & 48.1  & 61.9  & 19.7  & 29.5  & 26.6 \\
        \quad \textit{w/o} Low Rank & 0.0  & 0.0  & 0.4  & 4.4  & 9.1  & 2.9  & 2.0  & 2.7 \\
        \quad \textit{w/o} Spectral Normalize & 10.8  & 6.7  & 5.0  & 48.1  & 61.9  & 20.0  & 28.0  & 25.8 \\
        \quad \textit{w/o} Adaptive LR & 10.0  & 5.0  & 4.2  & 39.4  & 60.5  & 17.4  & 25.9  & 23.2 \\
        AdamW & 7.9  & 4.2  & 5.0  & 37.8  & 59.2  & 15.5  & 25.7  & 22.2 \\
        \midrule
        \multicolumn{9}{c}{\textit{Qwen3-4B-Base}} \\
        Ares (Ours) & 25.4  & 17.5  & 20.0  & 65.0  & 73.4  & 28.2  & 40.3  & 38.5 \\
        \quad \textit{w/o} Rank Momentum & 20.8  & 19.2  & 17.5  & 64.1  & 71.9  & 27.5  & 38.7  & 37.1 \\
        \quad \textit{w/o} Dynamic Rank & 23.3  & 20.0  & 17.9  & 62.5  & 71.6  & 28.9  & 38.7  & 37.6 \\
        \quad \textit{w/o} Low Rank & 0.8  & 1.7  & 0.8  & 17.2  & 23.6  & 7.6  & 8.2  & 8.6 \\
        \quad \textit{w/o} Spectral Normalize & 14.2  & 11.3  & 10.0  & 60.0  & 73.7  & 26.7  & 39.6  & 33.6 \\
        \quad \textit{w/o} Adaptive LR & 20.8  & 18.8  & 18.8  & 65.0  & 73.3  & 26.6  & 39.4  & 37.5 \\
        AdamW & 21.3  & 16.3  & 13.8  & 60.9  & 71.9  & 27.4  & 39.1  & 35.8 \\
        \bottomrule
    \end{tabular}
    \label{tab:ablation_study_full}
\end{table*}

Table~\ref{tab:ablation_study_full} extends the ablation of the main text to Qwen3-4B-Base, and the conclusions are consistent across scales. The most striking result is the collapse of \textit{w/o Low Rank}: removing the low-rank projection and updating in the full space drops the average accuracy to $2.7$ on Qwen3-1.7B-Base and $8.6$ on Qwen3-4B-Base, far below even AdamW. As noted in the main text, this row keeps the other Ares components (scale $s$, spectral normalization, and learning rate) fixed at values tuned for a low-rank gradient rather than re-tuning them for a full-rank update, so it is a controlled removal rather than an AdamW-equivalent baseline; the collapse indicates that the components are jointly calibrated around the low-rank gradient rather than that any single full-rank optimizer is unusable. Read together with the separately tuned AdamW baseline ($22.2$/$35.8$), it confirms the central premise of Sec.~\ref{sec:intro}: RLVR gradients carry their useful signal in a low-dimensional subspace, and confining the update to that subspace is what makes the optimizer work rather than an incidental efficiency trick. The remaining components each contribute a smaller but consistent gain: dynamic rank, spectral normalization, adaptive learning rate, and momentum are all beneficial at both scales, and notably \textit{w/o Dynamic Rank} already trails the full model, indicating that inferring the rank from the gradient spectrum, rather than fixing it, is responsible for part of the improvement.

\subsection{Detailed Results of Various Rank Adaptation Methods}
\label{app:rank_adaptation}

\begin{table*}[t]
    \centering
    \small
    \setlength{\tabcolsep}{5.5pt}
    \caption{Accuracy of Qwen models trained via GRPO with various optimizers on mathematical tasks. ``Avg.'' refers to average accuracy. The best is in bold and the second is underlined.}
      \begin{tabular}{lcccccccc}
        \toprule
        \textbf{Optimizer} & \textbf{AIME24} & \textbf{AIME25} & \textbf{AIME26} & \textbf{AMC23} & \textbf{MATH500} & \textbf{Minerva} & \textbf{OlyBench} & \textbf{Avg.} \\
        \midrule
        \multicolumn{9}{c}{\textit{Qwen3-1.7B-Base}} \\
        Exp & 15.0  & 10.8  & 10.0  & 43.4  & 61.6  & 20.4  & 29.4  & 27.2 \\
        Linear & 10.8  & 7.5  & 5.8  & 41.9  & 57.9  & 18.2  & 26.6  & 24.1 \\
        Threshold & 7.1  & 3.8  & 4.6  & 39.1  & 59.0  & 16.8  & 25.8  & 22.3 \\
        \midrule
        \multicolumn{9}{c}{\textit{Qwen3-4B-Base}} \\
        Exp & 25.4  & 17.5  & 20.0  & 65.0  & 73.4  & 28.2  & 40.3  & 38.5 \\      Linear & 24.6  & 24.6  & 17.1  & 64.4  & 72.5  & 28.4  & 40.2  & 38.8 \\
        Threshold & 20.4  & 15.0  & 15.0  & 56.3  & 71.5  & 25.9  & 37.9  & 34.6 \\
        \bottomrule
    \end{tabular}
    \label{tab:rank_adaptation_full}
\end{table*}

Table~\ref{tab:rank_adaptation_full} reports the full results behind Fig.~\ref{fig:rank_adaptation}, comparing three strategies for translating the estimated spectral energy into an update rank: an exponential mapping (Exp), a linear mapping (Linear), and a hard cutoff (Threshold). The smooth mappings clearly outperform the hard cutoff on both scales: Exp reaches $27.2$ / $38.5$ and Linear $24.1$ / $38.8$ average accuracy, whereas Threshold trails at $22.3$ / $34.6$. The Threshold strategy selects the rank by an abrupt energy cutoff and thus responds discontinuously to the gradient spectrum, discarding components near the boundary and destabilizing the update subspace as the spectrum evolves. In contrast, the smooth mappings let the rank track the concentration of gradient energy continuously over training, which is precisely the behavior motivated in Sec.~\ref{sec:intro}: the effective dimensionality of the update should follow the observed spectrum rather than jump at a fixed threshold.

\subsection{Performance on Various Domain Tasks}

\begin{table*}[t]
    \centering
    \small
    \caption{Accuracy of LLMs at different scales trained through different optimizers via GRPO on the MMLU-Pro sub-benchmarks. The best is in bold and the second best is underlined.}
    \setlength{\tabcolsep}{3pt}
      \begin{tabular}{lcccccccc}
      \toprule
       \textbf{Methods} & \textbf{Enginneering} & \textbf{Philosophy} & \textbf{CS} & \textbf{Physics} & \textbf{Math} & \textbf{Economics} & \textbf{Health} & \textbf{Avg.} \\
      \midrule
      \multicolumn{9}{c}{\textit{Qwen3-1.7B-Base}} \\
        AdamW & 25.6  & 29.3  & 35.0  & 26.6  & 27.5  & 48.9  & 34.8 & \underline{32.5} \\
        GaLore & 27.6  & 27.5  & 36.0  & 25.5  & 26.6  & 47.7  & 35.8 & 32.4 \\
        Ares (Ours) & 37.3  & 28.7  & 38.7  & 26.9  & 27.5  & 48.9  & 36.2 & \textbf{34.9} \\
      \midrule
      \multicolumn{9}{c}{\textit{Qwen3-4B-Base}} \\
        AdamW & 38.7  & 40.3  & 54.9  & 56.7  & 65.4  & 65.6  & 52.6 & \underline{53.5} \\
        GaLore & 37.9  & 41.7  & 53.2  & 55.0  & 64.5  & 64.3  & 53.4 & 52.9 \\
        Ares (Ours) & 46.9  & 45.5  & 58.1  & 62.4  & 71.1  & 69.1  & 59.2 & \textbf{58.9} \\
      \midrule
      \midrule
      \textbf{Methods} & \textbf{Other} & \textbf{History} & \textbf{Chemistry} & \textbf{Biology} & \textbf{Psychology} & \textbf{Law} & \textbf{Business} & \textbf{Avg.} \\
      \midrule
      \multicolumn{9}{c}{\textit{Qwen3-1.7B-Base}} \\
        AdamW & 32.5  & 28.6  & 25.8  & 64.5  & 53.6  & 21.0  & 30.3  & \underline{36.6} \\
        GaLore & 30.1  & 30.4  & 23.1  & 63.7  & 53.4  & 21.4  & 28.3  & 35.8 \\
        Ares (Ours) & 31.9  & 29.4  & 26.0  & 66.3  & 52.0  & 21.3  & 33.6  & \textbf{37.2} \\
      \midrule
      \multicolumn{9}{c}{\textit{Qwen3-4B-Base}} \\
        AdamW & 49.0  & 38.3  & 53.6  & 73.9  & 62.0  & 26.1  & 61.1  & \underline{52.0} \\
        GaLore & 45.3  & 40.4  & 52.2  & 71.5  & 61.5  & 26.7  & 61.7  & 51.3 \\
        Ares (Ours) & 49.8  & 43.8  & 61.4  & 77.4  & 65.8  & 29.5  & 65.7  & \textbf{56.2} \\
      \bottomrule
      \end{tabular}
      \label{tab:mmlu_pro}
\end{table*}

\begin{wrapfigure}{r}{0.4\linewidth}
    \includegraphics[width=\linewidth]{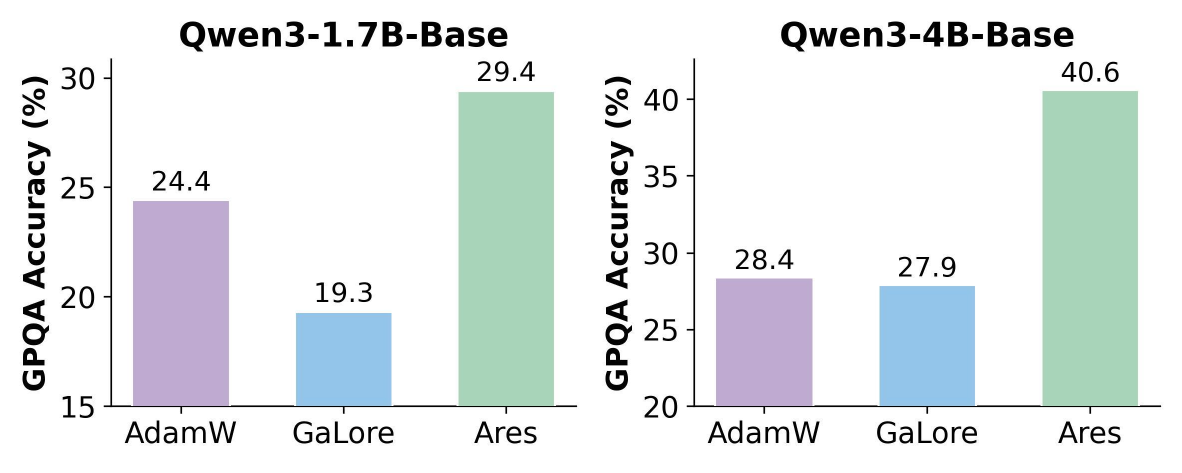}
    \caption{Accuracy of LLMs at different scales trained through different optimizers via GRPO on the GPQA.}
    \label{fig:gpqa}
\end{wrapfigure}

Although Ares is trained only on mathematical data, we assess whether its advantage transfers beyond mathematics. Table~\ref{tab:mmlu_pro} reports accuracy on the fourteen MMLU-Pro sub-benchmarks and Fig.~\ref{fig:gpqa} on GPQA. Ares achieves the best average accuracy across the board, improving over both AdamW and GaLore not only on STEM-heavy domains such as engineering, physics, and chemistry but also on distant fields such as philosophy, law, and business, and the margin widens at the 4B scale. The gains on GPQA follow the same pattern. This indicates that the low-rank spectral structure Ares exploits is a general property of RLVR training rather than an artifact of the mathematical reward, so optimizing within the RLVR gradient geometry yields improvements that generalize across domains.

\subsection{Effectiveness of Momentum}

\begin{figure*}[t]
    \includegraphics[width=\linewidth]{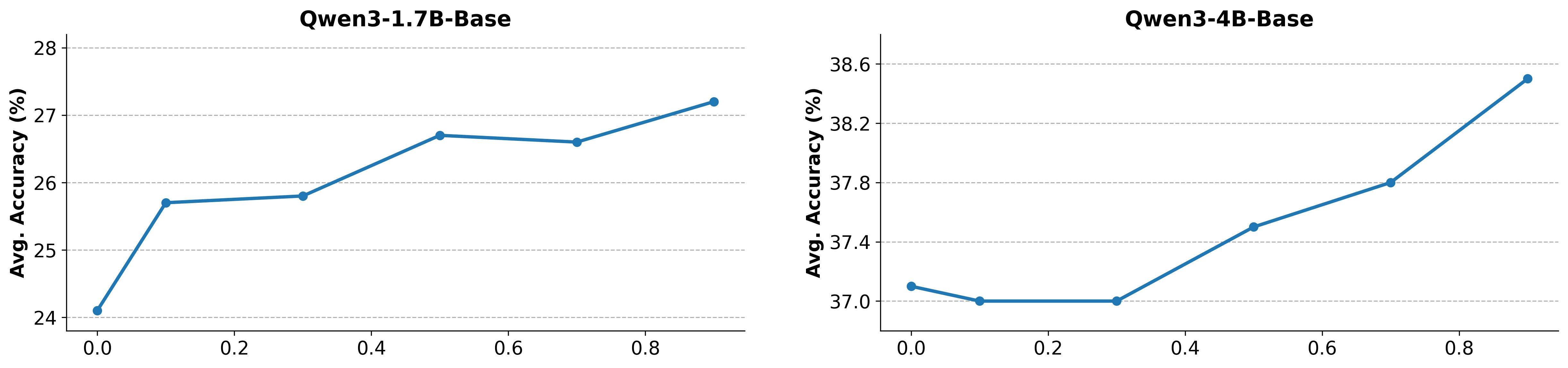}
    \caption{The effectiveness of different momentum coefficients.}
    \label{fig:momentum_effect}
\end{figure*}

Fig.~\ref{fig:momentum_effect} studies the sensitivity of Ares to the rank momentum coefficient $c$, which controls the exponential moving average that smooths the per-step effective rank $\hat r_t$ into the tracked rank $K$ (Sec.~\ref{sec:rank_estimation}). Overall, momentum is beneficial and its benefit is stable: it improves performance over the momentum-free variant, consistent with the \textit{w/o Rank Momentum} ablation in Table~\ref{tab:ablation_study_full}, and it delivers this gain without careful tuning, as accuracy remains consistently strong across a broad range of coefficients rather than depending on a narrowly tuned value.

The reason for this behavior lies in how $c$ governs the evolution of the tracked rank $K$. Because RLVR gradients are estimated from stochastic on-policy rollouts, the per-step estimate $\hat r_t$ is noisy, and $c$ determines how much of this noise is passed into $K$. A larger coefficient smooths $\hat r_t$ more strongly, so $K$ evolves gently and stably across steps; the update subspace then follows the slowly drifting spectrum without abrupt jumps, which yields the performance gains observed in the high-$c$ regime. A smaller coefficient, in contrast, lets the per-step noise pass through, so $K$ fluctuates sharply from step to step. These abrupt changes in the update dimensionality destabilize training and degrade performance. Momentum is thus most helpful when it is large enough to keep the rank trajectory smooth, and we adopt a coefficient in this stable regime rather than tuning it precisely.

\subsection{Sensitivity Analysis of Initial Rank}

\begin{figure*}[t]
    \includegraphics[width=\linewidth]{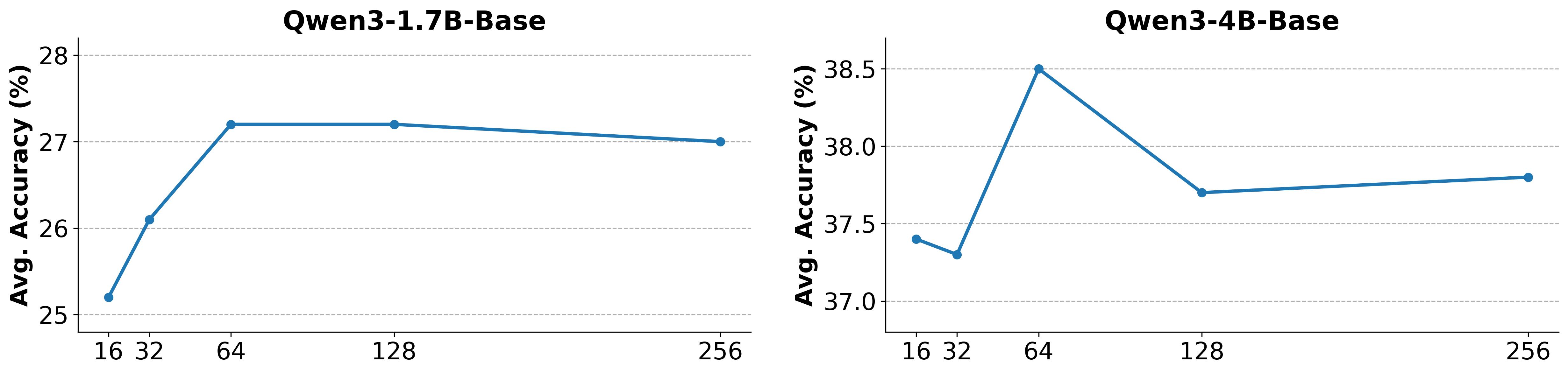}
    \caption{The effectiveness of various initial ranks.}
    \label{fig:init_rank_effect}
\end{figure*}

Fig.~\ref{fig:init_rank_effect} varies the initial rank $k$, which sets the target-rank budget within which Ares infers and tracks the effective rank (Sec.~\ref{sec:rank_estimation}). The behavior is governed by how well this budget covers the gradient energy. When $k$ is small, the retained leading subspace cannot capture the vast majority of the spectral energy, so the truncated reconstruction discards non-negligible components and performance is poor. Once $k$ is large enough to cover most of the energy, the picture changes qualitatively: the budget already contains the informative directions, and Ares is free to select the effective rank from within it. In this regime training is stable and accuracy is consistently strong across a wide range of $k$, in sharp contrast to fixed-rank methods whose accuracy hinges on the chosen $r$.

This behavior is exactly what the motivation of Sec.~\ref{sec:intro} predicts: a small rank already suffices to cover the vast majority of the gradient energy (Fig.~\ref{fig:rank_for_coverage}), so once the budget $k$ clears this modest threshold, enlarging it further adds only low-energy directions that Ares adaptively truncates away, leaving performance unchanged. The practical implication is that $k$ needs no careful tuning: one can simply set it somewhat large to safely cover the leading energy, and because the extra directions carry little energy and are discarded online, doing so incurs no meaningful cost. Ares thus removes the brittle rank-tuning burden of static low-rank optimizers, reducing the choice of $k$ from a sensitive hyperparameter to a loose upper bound.

\subsection{Adaptation on Various RLVR Algorithms}

\begin{table*}[t]
    \centering
    \small
    \setlength{\tabcolsep}{2pt}
    \caption{Effect of integrating Ares into various RLVR algorithms on mathematical tasks.}
      \begin{tabular}{lcccccccc}
        \toprule
         & \textbf{AIME24} & \textbf{AIME25} & \textbf{AIME26} & \textbf{AMC23} & \textbf{MATH500} & \textbf{Minerva} & \textbf{OlyBench} & \textbf{Avg.} \\
        \midrule
        \multicolumn{9}{c}{\textit{Qwen3-1.7B-Base}} \\
        RLOO \textit{w/} AdamW & 10.4  & 5.0  & 4.6  & 39.4  & 60.1  & 17.6  & 26.5  & 23.4 \\
        RLOO \textit{w/} Ares & 12.5  & 11.7  & 8.8  & 48.8  & 61.8  & 19.7  & 29.6  & 27.6 \\
        $\Delta$ & \textcolor{darkgreen}{+2.1}  & \textcolor{darkgreen}{+6.7}  & \textcolor{darkgreen}{+4.2}  & \textcolor{darkgreen}{+9.4}  & \textcolor{darkgreen}{+1.7}  & \textcolor{darkgreen}{+2.1}  & \textcolor{darkgreen}{+3.1}  & \textcolor{darkgreen}{+4.2} \\
        \midrule
        REINFORCE++ \textit{w/} AdamW & 7.5  & 2.1  & 2.9  & 36.7  & 61.4  & 17.0  & 26.4  & 22.0 \\
        REINFORCE++ \textit{w/} Ares & 10.0  & 6.3  & 4.6  & 43.4  & 61.5  & 19.3  & 29.0  & 24.9 \\
        $\Delta$ & \textcolor{darkgreen}{+2.5}  & \textcolor{darkgreen}{+4.2}  & \textcolor{darkgreen}{+1.7}  & \textcolor{darkgreen}{+6.7}  & \textcolor{darkgreen}{+0.1}  & \textcolor{darkgreen}{+2.3}  & \textcolor{darkgreen}{+2.6}  & \textcolor{darkgreen}{+2.9}   \\
        \midrule
        GDPO \textit{w/} AdamW & 6.3  & 4.6  & 4.6  & 40.0  & 59.3  & 15.3  & 25.2  & 22.2 \\
        GDPO \textit{w/} Ares & 13.8  & 8.8  & 9.6  & 46.9  & 62.6  & 19.3  & 29.7  & 27.2 \\
        $\Delta$ &  \textcolor{darkgreen}{+7.5}  & \textcolor{darkgreen}{+4.2}  & \textcolor{darkgreen}{+5.0}  & \textcolor{darkgreen}{+6.9}  & \textcolor{darkgreen}{+3.3}  & \textcolor{darkgreen}{+4.0}  & \textcolor{darkgreen}{+4.5}  & \textcolor{darkgreen}{+5.1}  \\
        \midrule
        \midrule
        \multicolumn{9}{c}{\textit{Qwen3-4B-Base}} \\
        RLOO \textit{w/} AdamW & 19.2  & 15.8  & 14.1  & 64.4  & 71.1  & 26.7  & 38.3  & 35.7 \\
        RLOO \textit{w/} Ares & 24.6  & 20.4  & 22.5  & 64.1  & 73.6  & 28.6  & 40.7  & 39.2 \\
        $\Delta$ & \textcolor{darkgreen}{+5.4}  & \textcolor{darkgreen}{+4.6}  & \textcolor{darkgreen}{+8.4}  & \textcolor{red}{-0.3} & \textcolor{darkgreen}{+2.5}  & \textcolor{darkgreen}{+1.9}  & \textcolor{darkgreen}{+2.4}  & \textcolor{darkgreen}{+3.6}   \\
        \midrule
        REINFORCE++ \textit{w/} AdamW & 13.8  & 11.2  & 7.9  & 53.1  & 71.7  & 26.1  & 36.7  & 31.5 \\
        REINFORCE++ \textit{w/} Ares & 18.3  & 15.8  & 18.3  & 64.1  & 73.4  & 26.7  & 37.6  & 36.3 \\
        $\Delta$ &  \textcolor{darkgreen}{+4.5}  & \textcolor{darkgreen}{+4.6}  & \textcolor{darkgreen}{+10.4}  & \textcolor{darkgreen}{+11.0}  & \textcolor{darkgreen}{+1.7}  & \textcolor{darkgreen}{+0.6}  & \textcolor{darkgreen}{+0.9}  & \textcolor{darkgreen}{+4.8}  \\
        \midrule
        GDPO \textit{w/} AdamW & 21.6  & 16.7  & 16.7  & 59.7  & 72.2  & 26.7  & 39.2  & 36.1 \\
        GDPO \textit{w/} Ares & 22.9  & 18.8  & 15.4  & 60.9  & 71.4  & 28.1  & 40.2  & 36.8 \\
        $\Delta$ &  \textcolor{darkgreen}{+1.3}  & \textcolor{darkgreen}{+2.1}  & \textcolor{red}{-1.3} & \textcolor{darkgreen}{+1.2}  & \textcolor{red}{-0.8} & \textcolor{darkgreen}{+1.4}  & \textcolor{darkgreen}{+1.0}  & \textcolor{darkgreen}{+0.7}  \\
        \bottomrule
    \end{tabular}
    \label{tab:other_rlvr_algorithm}
\end{table*}

Ares is defined at the level of the optimizer and is therefore agnostic to the choice of RLVR objective. To verify this, Table~\ref{tab:other_rlvr_algorithm} replaces AdamW with Ares under three additional algorithms, \ie RLOO~\citep{ahmadian2024rloo}, REINFORCE++~\citep{hu2025reinforceplusplus}, and GDPO~\citep{liu2026gdpo}, beyond the GRPO setting of the main results. Ares improves the average accuracy for every algorithm at both scales, by up to $+5.1$ points, with only a handful of per-benchmark regressions. The consistency of these gains reinforces the analysis in Sec.~\ref{sec:intro}: the concentration of gradient energy is a property of RLVR training in general, driven by the verifiable reward signal rather than by any particular advantage estimator, so an optimizer that adapts to this shared spectral structure benefits across RLVR algorithms.

\subsection{Effectiveness of Ares on SFT Process}

\begin{figure}[t]
    \includegraphics[width=\linewidth]{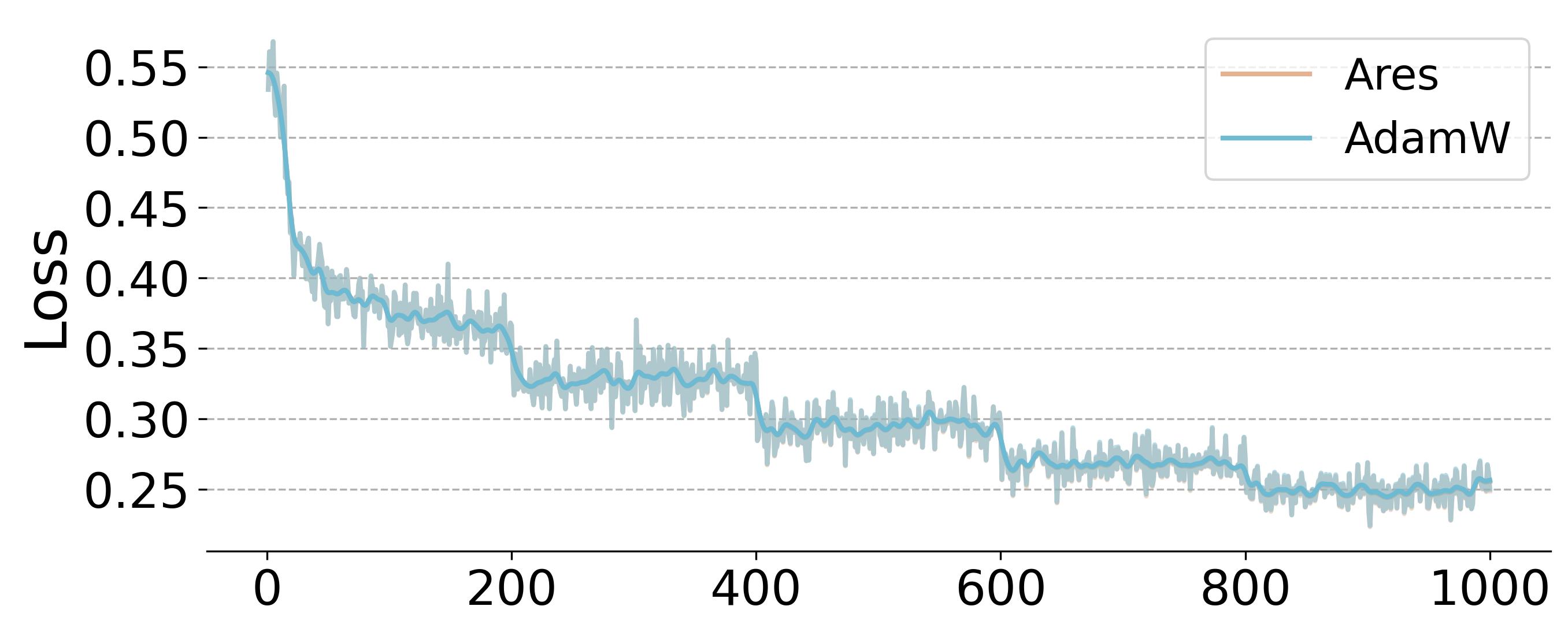}
    \caption{Loss changes during SFT training with different optimizers.}
    \label{fig:sft_loss}
\end{figure}

\begin{table*}[t]
    \centering
    \small
    \caption{Performance of Qwen3-4B-Base through SFT training with different optimizers.}
      \begin{tabular}{lcccccccc}
        \toprule
         & \textbf{AIME24} & \textbf{AIME25} & \textbf{AIME26} & \textbf{AMC23} & \textbf{MATH500} & \textbf{Minerva} & \textbf{OlyBench} & \textbf{Avg.} \\
        \midrule
        AdamW & 17.0  & 21.0  & 17.3  & 58.3  & 69.2  & 26.5  & 34.3  & 34.8 \\
        Ares  & 19.9  & 20.3  & 18.1  & 58.3  & 69.5  & 26.7  & 34.3  & 35.3 \\
        \bottomrule
    \end{tabular}
    \label{tab:sft_performance}
\end{table*}

Finally, we test whether the advantage of Ares is specific to RLVR by applying it to supervised fine-tuning (SFT). Fig.~\ref{fig:sft_loss} shows that Ares and AdamW follow nearly identical loss curves, and Table~\ref{tab:sft_performance} shows that their downstream accuracies are on par ($35.3$ vs.\ $34.8$ average), with no consistent gain. This is exactly the outcome anticipated in Sec.~\ref{sec:intro}: SFT gradients are far less spectrally concentrated than those of RLVR, so there is little low-rank structure for Ares to exploit and it reduces to a standard adaptive optimizer. The contrast is the strongest evidence for our central thesis, \ie the benefit of Ares comes not from the low-rank machinery perse but from matching the optimizer to the distinctive gradient geometry of RLVR. Optimizers should be designed for the geometry of the gradients they actually see.

\end{document}